\documentclass[10pt,twocolumn,letterpaper]{article}

\usepackage{iccv}
\usepackage{times}
\usepackage{epsfig}
\usepackage{graphicx}
\usepackage{amsmath}
\usepackage{amssymb}
\usepackage{tikz}

\usepackage[pagebackref=true,breaklinks=true,colorlinks,bookmarks=false]{hyperref}

\iccvfinalcopy 

\ificcvfinal\fi

\usepackage{cleveref}
\usepackage{wrapfig}
\usepackage{color}
\usepackage{subcaption}
\usepackage{booktabs}
\usepackage{makecell}
\usepackage{multirow}
\usepackage{enumitem}
\usepackage{algpseudocode}
\usepackage{algorithm}
\usepackage{algorithmicx}  
\usepackage{nicefrac}
\usepackage{tikz}

\usepackage[toc,page]{appendix}

\newlength{\xxx}
\newsavebox{\measurebox}

\newcommand*\Let[2]{\State #1 $\gets$ #2}  
\newcommand*\Sample[2]{\State #1 $\sim$ #2} 
\algrenewcommand\algorithmicrequire{\textbf{Precondition:}}  
\algrenewcommand\algorithmicensure{\textbf{Postcondition:}}

\def\checkmark{\tikz\fill[scale=0.3](0,.35) -- (.25,0) -- (1,.7) -- (.25,.15) -- cycle;}

\newcommand{\Xc}{\mathcal{X}}
\newcommand{\Zc}{\mathcal{Z}}
\newcommand{\Yc}{\mathcal{Y}}

\makeatletter
\newcommand{\printfnsymbol}[1]{%
  \textsuperscript{\@fnsymbol{#1}}%
}
\makeatother

\newcommand{\modelname}{MG-GAN}
\newcommand{\probnetname}{PM-Net}

\newcommand{\norm}[1]{\left\lVert#1\right\rVert}

\newcommand{\LSTM}{\text{LSTM}}
\newcommand{\ATT}{\text{ATT}}

\begin{document}

\title{MG-GAN: A Multi-Generator Model Preventing Out-of-Distribution Samples in Pedestrian Trajectory Prediction}

\author{
Patrick Dendorfer\thanks{Equal contribution.} \hspace{1cm} Sven Elflein\printfnsymbol{1} \hspace{1cm} Laura Leal-Taix\'e \vspace*{1mm}\\
 Technical University Munich \\
{\tt\small \{patrick.dendorfer,sven.elflein,leal.taixe\}@tum.de}
}

\maketitle


\thispagestyle{plain}
\pagestyle{plain}
\begin{abstract}
Pedestrian trajectory prediction is challenging due to its uncertain and multimodal nature. %
While generative adversarial networks can learn a distribution over future trajectories, they tend to predict out-of-distribution samples when the distribution of future trajectories is a mixture of multiple, possibly disconnected modes. 
To address this issue, we propose a multi-generator model for pedestrian trajectory prediction. 
Each generator specializes in learning a distribution over trajectories routing towards one of the primary modes in the scene, while a second network learns a categorical distribution over these generators, conditioned on the dynamics and scene input. This architecture allows us to effectively sample from specialized generators and to significantly reduce the out-of-distribution samples compared to single generator methods.

\end{abstract}

\section{Introduction}
\label{sec:introduction}
\begin{figure*}
      \begin{subfigure}[t]{0.32\textwidth}
     \centering
        \includegraphics[width=0.8\textwidth]{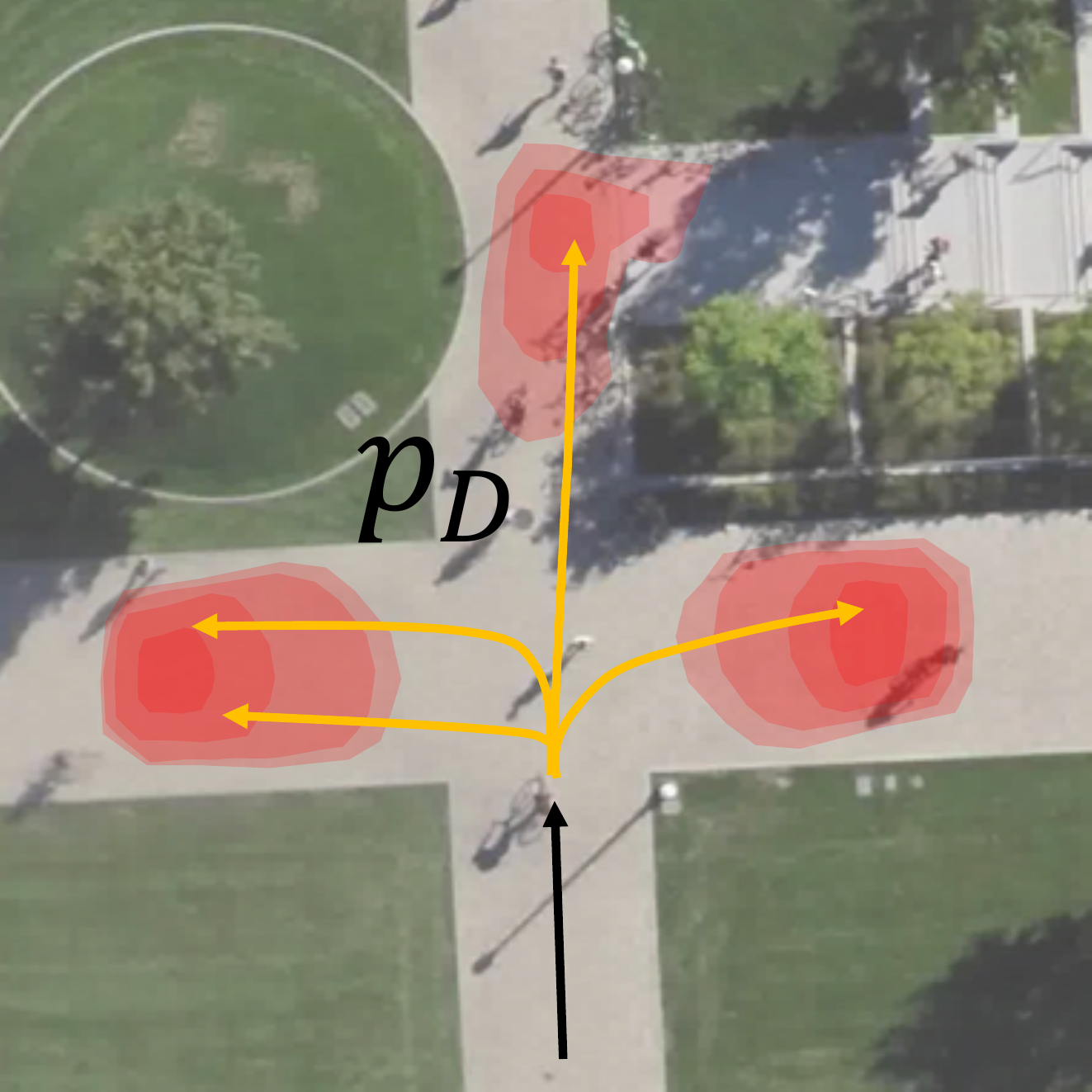}
        \caption{Target Distribution $p_D$}
        \label{fig:teaser:GT}
    \end{subfigure}
    \hfill
    \begin{subfigure}[t]{0.32\textwidth}
    \centering
    \includegraphics[width=0.8\textwidth]{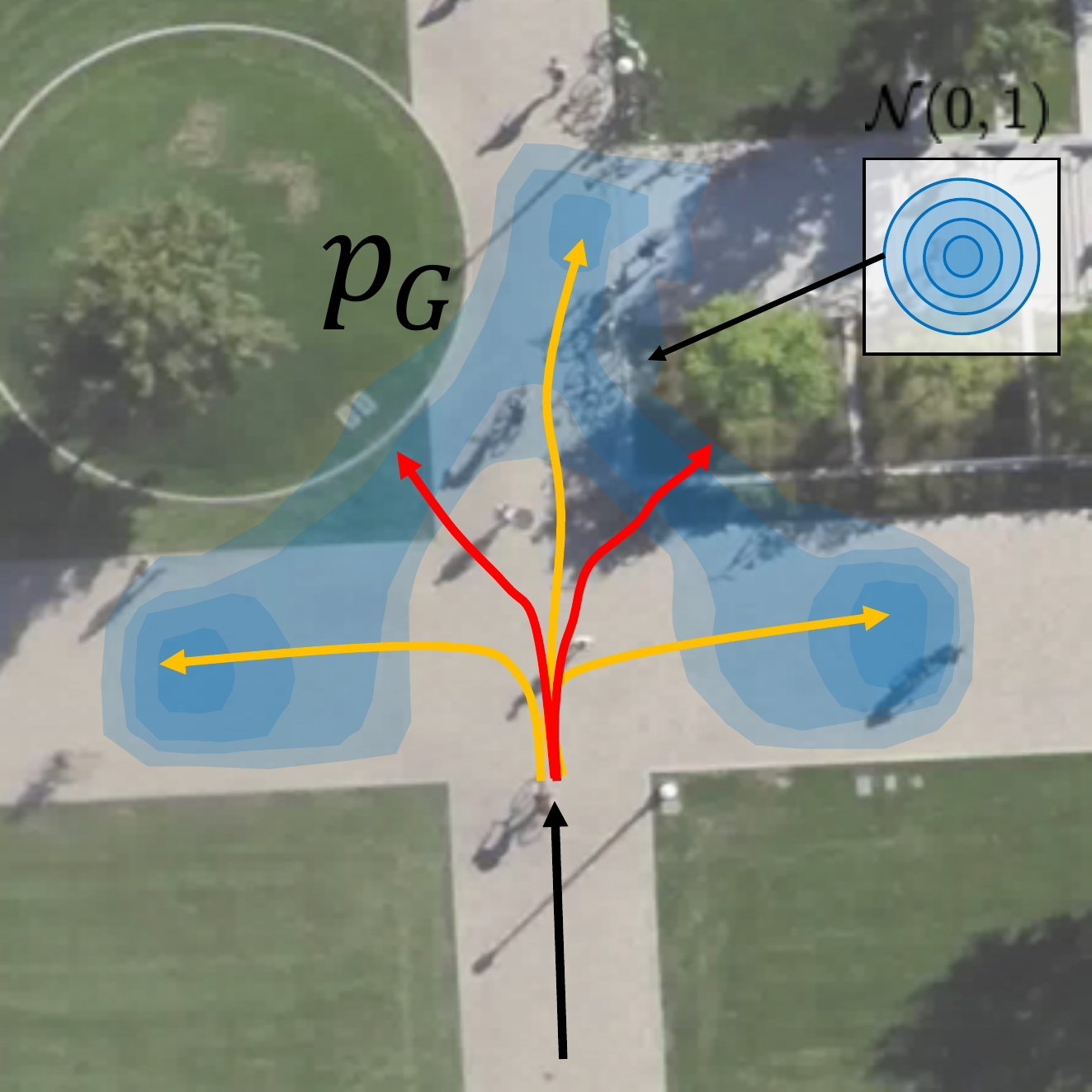}
    \caption{Single Generator distribution $p_G$}
    \label{fig:teaser:GAN}
    \end{subfigure}
    \hfill
      \begin{subfigure}[t]{0.32\textwidth}
     \centering
        \includegraphics[width=0.8\textwidth]{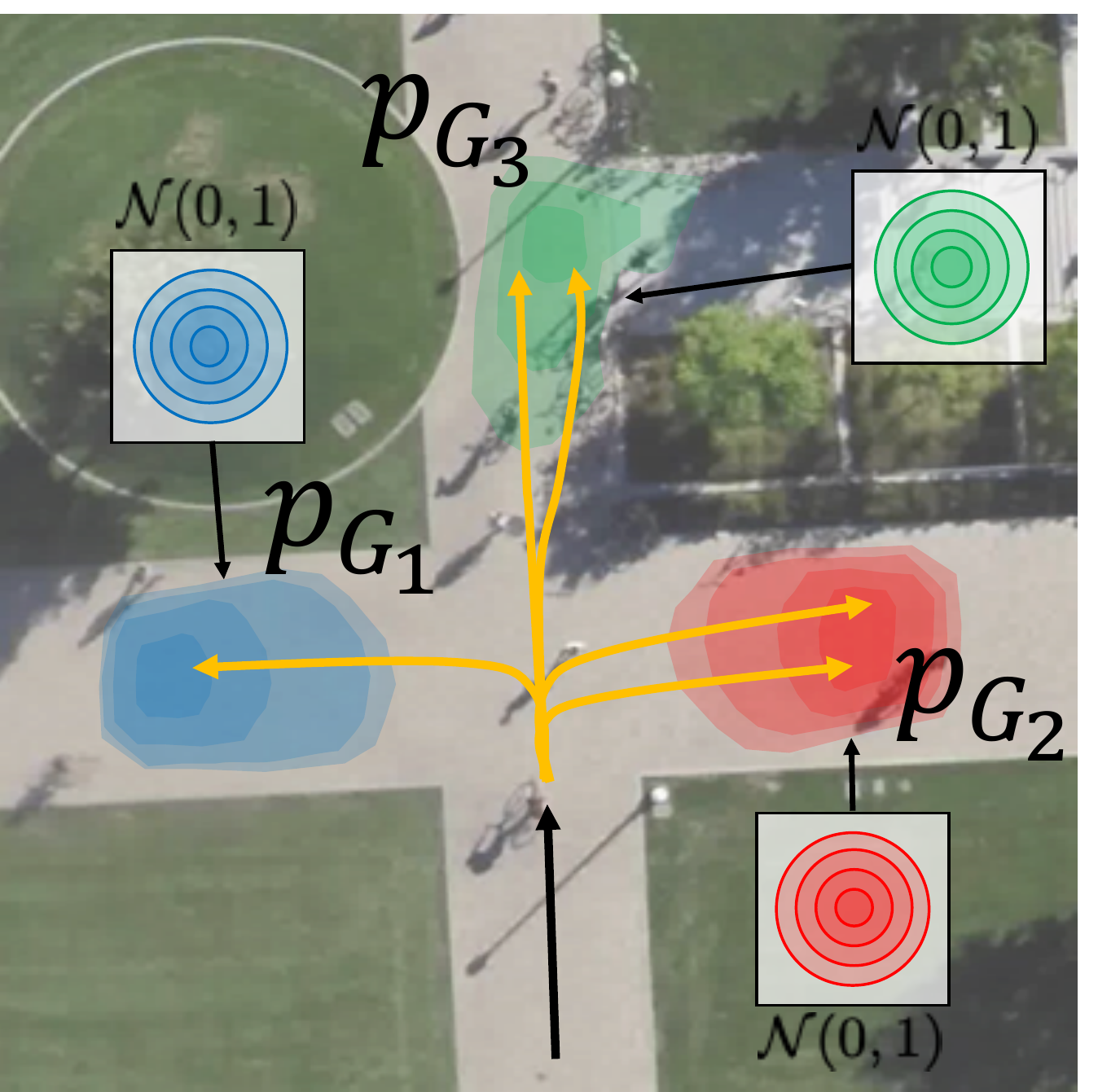}
        \caption{Multi-generator distribution}
        \label{fig:teaser:MGAN}
    \end{subfigure}
    
  \caption{The figure illustrates a pedestrian reaching a junction (black) including (a)  the multimodal target distribution of future paths, (b) learned future trajectory distribution by a single generator GAN predicting out-of-distribution samples (red), and (c) learned trajectory distribution of multi-generator mixture model. }
\label{fig:teaser}
\end{figure*}

To safely navigate through crowded scenes, intelligent agents such as autonomous vehicles or social robots need to anticipate human motion. Predicting human trajectories is particularly difficult because future actions are multimodal: given a past trajectory, there exist several plausible future paths, depending on the scene layout and social interactions among pedestrians. 
Recent methods leverage conditional generative adversarial networks (GANs)~\cite{gans,social_gan, sadeghian2018sophie, bigat} to learn a distribution over trajectories. These methods present significant improvements over deterministic models~\cite{sociallstm,social_force}. However, they suffer from limitations observed in the context of GANs~\cite{noGANlandICML, disconnectedManifold} that manifest in mode collapse or prediction of undesired out-of-distribution (OOD) samples, effectively yielding non-realistic trajectories.
Mode collapse can be tackled with best-of-many sampling~\cite{best-of-many-sampling} or regularizations of the latent space~\cite{bigat, social_ways} but the problem of OOD samples remains unsolved. 
These OOD samples are particularly problematic in real-world applications where high precision of predictions matters.
Imagine an autonomous vehicle driving through crowded environments and interacting with pedestrians. To ensure the safety of pedestrians, the vehicle needs to anticipate their future motion and react accordingly, \eg, brake or turn. As a consequence, unrealistic predictions may lead to sudden reactions that pose danger to other traffic participants.

To understand why OOD samples are produced by state-of-the-art GAN methods, we need to understand the underlying geometry of the problem.
Consider a pedestrian reaching the junction in \Cref{fig:teaser:GT}. There are three plausible main directions that the pedestrian can take, namely, going straight, left, or right. Furthermore, there exist several paths that route towards these directions. 
While all recent works agree that such trajectory distribution is inherently multimodal, we further observe that the distribution consists of several disconnected modes. Each mode is shown in \Cref{fig:teaser:MGAN} in different colors, and as we can observe, the three modes are disconnected in space. 
Existing GAN models do not consider this property, and hence generate undesirable OOD samples in between modes, visualized as red trajectories in \Cref{fig:teaser:GAN}. 
This is an inherent problem of single-generator GANs, as they cannot learn a mapping from a continuous latent space to a disconnected, multimodal target distribution~\cite{noGANlandICML}.   

In this paper, we address this issue and explicitly focus on learning such disconnected multimodal distributions for pedestrian trajectory prediction. 
To this end, we propose a novel multi-generator GAN that treats the multimodal target distribution as a mixture of multiple continuous trajectory distributions by optimizing a continuous generator for each mode. 
Unlike previous multi-generator models~\cite{mgan, infogan}, our model needs to adapt to the selection of generators to different scenes, \eg, two- and three-way junctions. For this, we employ a fixed number of generators and allow the model to learn the necessary number of modes directly from visual scene information.
Towards this end, we train a second module estimating the categorical probability distribution over the individual generators, conditioned on the input observations. At test time, we first select a specific generator based on its categorical probability and sample then trajectories specialized to that particular mode present in the scene.
For measuring the quality of the predictions, we extend the concept of traditional $L_2$ error measures with a precision and recall metric~\cite{FirstPrecisionRecallNIPS, ImprovedPrecisionRecallNIPS}. Our experimental evaluation shows that our proposed model overcomes state-of-the-art and single-generator methods when comparing the behavior of predicting OOD samples. 

We summarize our \textbf{main contributions} as follows: (i) we discuss the limitations of single generator GANs and propose a novel multi-generator method that learns a multimodal distribution over future trajectories, conditioned on the visual input. To this end, we (ii) present a model that estimates a conditional distribution over the generators and elaborate a training scheme that allows us to jointly train our model end-to-end. Finally, (iii) we introduce recall and precision metrics for pedestrian trajectory prediction to measure the quality of the entire predictive distribution, and in particular OOD samples. We demonstrate our method's efficiency and robustness through extensive ablations.
The source code of the model and experiments is available: \href{https://github.com/selflein/MG-GAN}{https://github.com/selflein/MG-GAN}.

\section{Related Work}
\label{sec:related_work}
\paragraph{ Trajectory Forecasting.}
Since its inception, the field of pedestrian trajectory prediction has moved from handcrafted~\cite{social_force} to data-driven~\cite{sociallstm} methods. While the first learning methods used deterministic LSTM encoder-decoder architectures (S-LSTM~\cite{sociallstm}), deep generative models~\cite{social_gan, sadeghian2018sophie, bigat, social_ways, fernando2018gdgan, dendorfer20accv} quickly emerged as state-of-the-art prediction methods. This development enabled the shift from predicting a single future trajectory to producing a distribution of possible future trajectories.
S-GAN~\cite{social_gan} establishes a conditional Generative Adversarial Networks~\cite{gans} to learn the ground-truth trajectory distribution and S-GAN-P~\cite{social_gan}and SoPhie~\cite{sadeghian2018sophie} extend S-GAN with visual and social interaction components. 
Further, S-BiGAT~\cite{bigat} increases the diversity of the samples by leveraging bicycle GAN training~\cite{zhu2018multimodal} that encourages the connection between the output and the latent code to be invertible. Goal-GAN~\cite{dendorfer20accv} circumvents the problem of mode collapse by conditioning the decoder on a goal position estimated based on the topology of the scene.

GANs~\cite{gans} have well-known issues with mode collapse, this is why many models~\cite{social_gan, sadeghian2018sophie} use an $L_2$ variety loss~\cite{best-of-many-sampling} or modify the GAN objective~\cite{social_ways} to encourage diversity of the samples. While producing highly diverse samples ensures coverage of all modes in the distribution, we also obtain many unrealistic out-of-distribution samples. 
The problem of OOD samples has been remained unnoticed partially due to the evaluation metrics used in the field which only measure the minimum $L_2$ distance between the set of predictions and the ground truth, namely the recall. Nonetheless, the realism of predicted trajectories, equivalent to a precision metric, is seldomly evaluated. We advocate that trajectory prediction methods should be evaluated concerning both of the aforementioned aspects.
 
Other work uses conditional variational autoencoders (VAE)~\cite{VariationalAutoencoder} for multimodal pedestrian trajectory prediction~\cite{desire,sadeghian2017carnet, li2019conditional, bhattacharyya2019conditional}. More recently, Trajectron++~\cite{Salzmann2020TrajectronDT} uses a VAE and represents agents' trajectories in a graph-structured recurrent neural network. PECNet~\cite{mangalam2020pecnet} proposes goal-conditioned trajectory forecasting. Similar to GANs, VAEs are also continuous transformations and suffer from the limitations of generating distributions on disconnected manifolds~\cite{rolfeDiscreteVariationalAutoencoders2017}. 

Lastly, P2TIRL~\cite{deo2020trajectory} learns a grid-based policy with
maximum entropy inverse reinforcement learning.
In summary, existing methods pay little attention to the resulting emergence of out-of-distribution samples and do not discuss the topological limitation in learning a distribution on disconnected supports.

\vspace{\xxx}
\paragraph{Generation of Disconnected Manifolds.}
Understanding the underlying geometry of the problem is important when training deep generative models~\cite{fefferman2013testing}. More precisely, learning disconnected manifolds requires disconnectedness within the model. A single generator preserves the topology of the continuous latent space and cannot exclusively predict samples on disconnected manifolds~\cite{noGANlandICML}. 

For image generation, the problem of multimodal learning is well-known and widely studied. Addressing this issue, \cite{noGANlandICML} proposes a rejection sampling method based on the norm of the generator’s Jacobian. InfoGAN~\cite{infogan} discretizes the latent space by introducing extra dimensions. Other works use mixtures of generators~\cite{gan_ensembles, adagan, zhong2019rethinking, probgan, MADGAN, MIXGAN} to construct a discontinuous function. However, these models assume either a uniform or unconditional probability for their discrete latent code or generators. As a result, these methods are unable to adapt to different scenes and thus unsuitable for the trajectory prediction task.

Our research is the first to address the problem of learning disconnected manifolds using multiple generators for the task of pedestrian trajectory prediction by modeling a conditional distribution over the generators.

\section{Problem Definition}
\label{sec:background}
In this work, we tackle the problem of jointly predicting future trajectories for all pedestrians in the scene. 
For each pedestrian $i$, we generate a set of $K$ future trajectories $\{\hat{Y}^k_i\}_{k=1, \dots, K}$ with $t \in \left[t_{obs}+ 1, t_{pred} \right]$ for a given input trajectory $X_i$ with $t \in \left[t_{1}, t_{obs} \right]$. 
This implies learning the true distribution of trajectories conditioned on the input trajectories and scene layout.

In many real-world scenarios such as in \Cref{fig:teaser}, the target distribution $p_D$ is multimodal and composed of disconnected modes. 
\vspace{\xxx}
\paragraph{Why do Single Generator GANs produce OOD Samples?}
State-of-the-art methods use the standard conditional GAN architecture~\cite{gans} and its modifications~\cite{Metz2017UnrolledGA, wassersteinGAN} to learn a distribution over future trajectories.
These models learn a continuous mapping $G: \Xc \times \Zc \to \mathcal{Y}$ from the latent space $\Zc$ combined with the observations' space $\Xc$ to the space of future trajectories $\Yc$.
The probability prior $p(z)$ on  $\Zc$ is mainly a standard multivariate normal distribution with $ z \sim \mathcal{N}(0,\,1)$. When modeling $G$ with a neural network, the mapping is continuous and preserves the topology of the space. Hence, the transformation $G\left(x,  \Zc \right)$ of the support of the probability distribution $\Zc$ is connected in the output space~\cite{noGANlandICML}.  
Therefore, theoretic work~\cite{noGANlandICML, disconnectedManifold} discusses that learning a distribution on disconnected manifolds is impossible; we also observe this phenomenon in our experiments.
\vspace{\xxx}
\paragraph{Why are OOD Samples problematic?}
Real world-applications relying on trajectory predictions, \eg autonomous vehicles, have to treat every prediction as a possible future scenario and need to adjust their actions accordingly. Thus, not only missed but also unrealistic predictions may crucially hurt the performance of those applications. As OOD samples without support in the ground-truth distribution are likely to be unrealistic, we aim to keep their number small while still covering all modes.
\vspace{\xxx}
\paragraph{How can we prevent OOD Samples?}
All single generator models will predict OOD if the target distribution lies on disconnected manifolds.
Theoretically, there are only two ways to achieve disconnectedness in $\Yc$: making $\Zc$ disconnected or making the generator mapping $G: \Xc \times \Zc \to \Yc$ discontinuous. We discuss both approaches in our paper but find the latter to be more effective. 

\vspace{\xxx}
\paragraph{How to measure OOD Samples?}
Best-of-many $L_2$ distance metrics focus on minimizing the error between a single sample out of a set of predictions without assessing the quality of the remaining trajectories. 
Therefore, we compare our model on both, recall and precision~\cite{FirstPrecisionRecallNIPS, ImprovedPrecisionRecallNIPS}, which are commonly used to assess the quality of generative models. While existing distance measures highly correlate with recall, we are equally interested in precision that correlates with the number of OOD samples.

\section{Method}
\label{sec:method}

\begin{figure*}
 \centering
    \includegraphics[width=\linewidth]{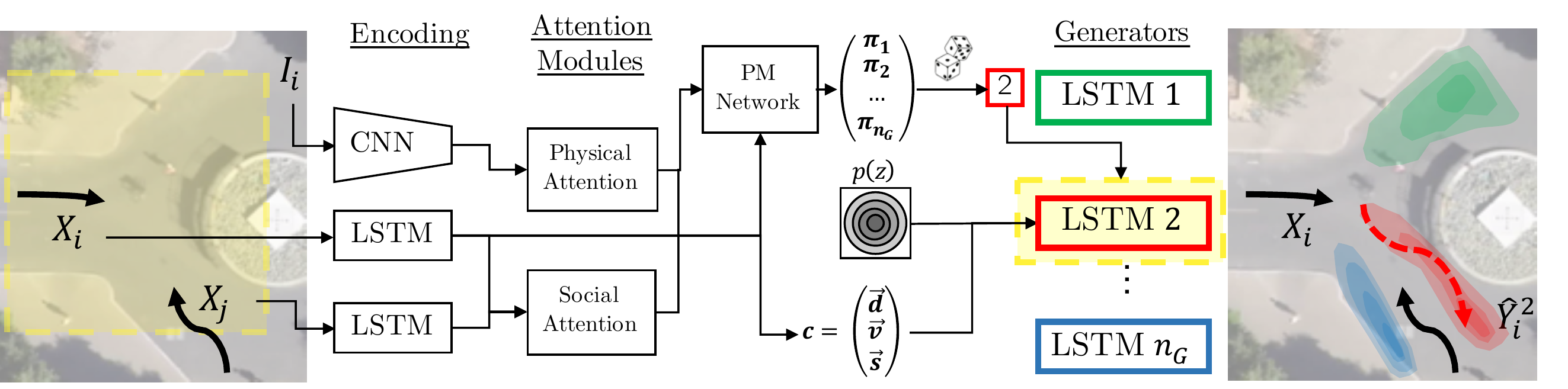}
    \caption{Architecture of \modelname{}. The scene image $I_i$ and observed trajectories $X$ are encoded and passed to the physical and social attention modules. The $n_G$ generators can predict different conditional trajectory distributions for the given scene observation. The \probnetname{} estimates probabilities $\boldsymbol{\pi}$ for the generators. The model samples or selects a generator from $\boldsymbol{\pi}$ and predicts a trajectory $\hat{Y}$ conditioned on the features $c$ and the noise vector $z$.}
    \label{fig:model_architecure}
\end{figure*}

In this section, we present our multi-generator framework for pedestrian trajectory prediction. Our model learns a discontinuous function as a mixture of distributions modeled by multiple generators (\Cref{sec:method:mgan}). 

To adapt to different scenes, we train a second network estimating the categorical distribution over generators (\Cref{sec:method:pinet}) for new unseen scenes. 

\subsection{\modelname{}}
\label{sec:method:mgan}
\paragraph{ Visual and Trajectory Encoders.}
We outline the architecture of our model in \Cref{fig:model_architecure}. First, the feature encoders extract visual and dynamic features $d_i$ from the input sequences $X_i$ and scene image patches $I_i$ of each pedestrian $i$. The attention modules combine these encodings to compute the physical attention~\cite{sadeghian2018sophie} features $v_i$ and social attention~\cite{social_ways} features $s_i$. 
After the encoding and attention, we concatenate the dynamic $d_i$, physical $v_i$, and social $s_i$ features to $c_i = \left[ d_i , v_i, s_i \right]$. In the following, we omit the index indicating individual pedestrians to avoid notation clutter. %
Note that we leverage established modules to model physical and social interactions~\cite{sadeghian2018sophie, social_ways, social_gan}, as our contribution is the multi-generator framework. We provide more details on these components in the supplementary. 

\vspace{\xxx}
\paragraph{Multi-generator Model.} 
In our model, we leverage $n_G$ different generators $\{G_g\}$, where each generator specializes in learning a different trajectory distribution conditioned on the input $c$.  All generators share the same network architecture, however, they do not share weights. 
The generator architecture consists of a LSTM decoder, initialized with the features $c$ and a random noise vector $ z \sim \mathcal{N}(0,\,1)$ as the initial hidden state $h^{0}$. 
The final trajectory $\hat{Y}$ is then predicted recurrently:
\begin{equation}
\Delta \hat{Y}^t = \LSTM_g\left( \Delta X^{t-1}, h^{t-1}\right).
\end{equation}

Existing multi-generator modules proposed in the context of image generation assume the distribution over the generators to be constant~\cite{mgan, probgan}. 
However, in the case of trajectory prediction, the number of modes is unknown a priori. Therefore, we train a module that adapts to the scene by activating specific generators, conditioned on the observations and interactions $c$.

\begin{figure*}
    \centering
    \begin{subfigure}[t]{0.18\linewidth}
        \includegraphics[width=\textwidth]{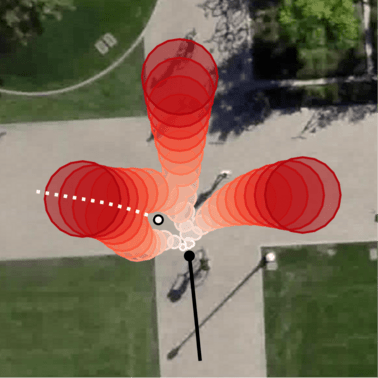}%
        \vfill
        \includegraphics[width=\textwidth]{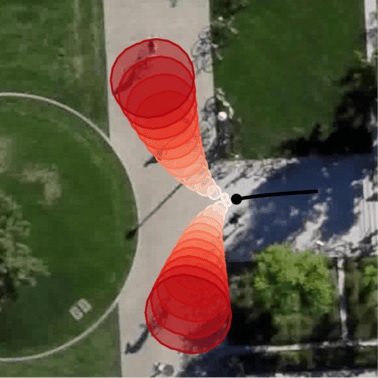}
        \caption{GT}
        \label{fig:synthetic:upperGT} 
    \end{subfigure}%
    \hfill
    \begin{subfigure}[t]{0.18\linewidth}
        \includegraphics[width=\textwidth]{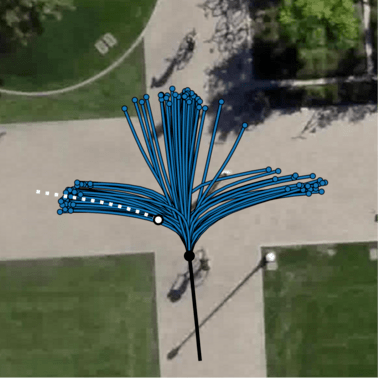}
        \vfill
        \includegraphics[width=\textwidth]{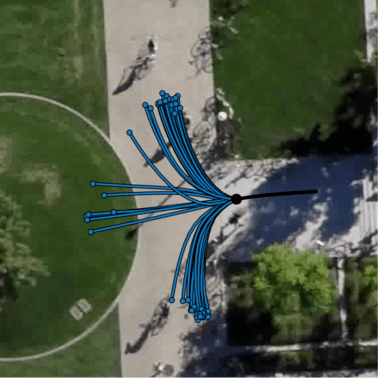}
        \caption{GAN L2}
        \label{fig:synthetic:upperGANL2}
    \end{subfigure}%
    \hfill
    \begin{subfigure}[t]{0.18\linewidth}
        \includegraphics[width=\textwidth]{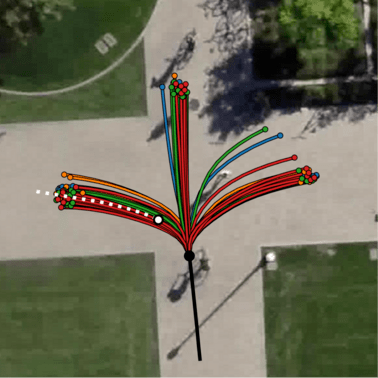}
        \vfill
        \includegraphics[width=\textwidth]{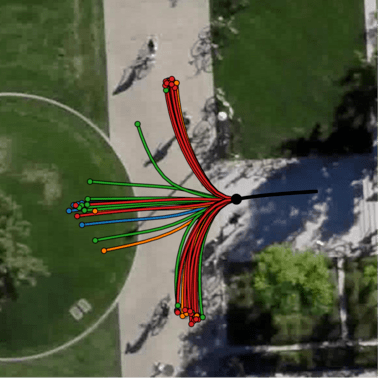}
        \caption{InfoGAN}
        \label{fig:synthetic:upperInfoGAN}
    \end{subfigure}%
    \hfill
    \begin{subfigure}[t]{0.18\linewidth}
        \includegraphics[width=\textwidth]{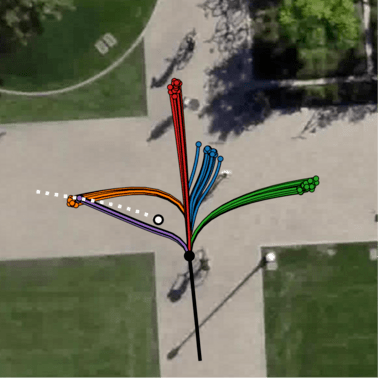}
        \includegraphics[width=\textwidth]{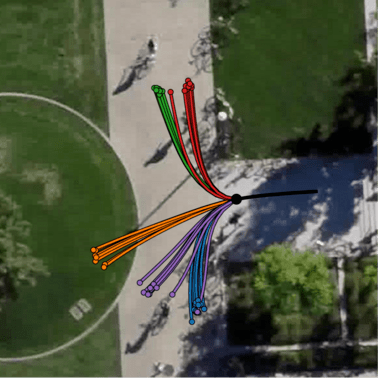}
        \caption{MGAN}
        \label{fig:synthetic:upperMGAN}
    \end{subfigure}%
    \hfill
    \begin{subfigure}[t]{0.18\linewidth}
        \includegraphics[width=\textwidth]{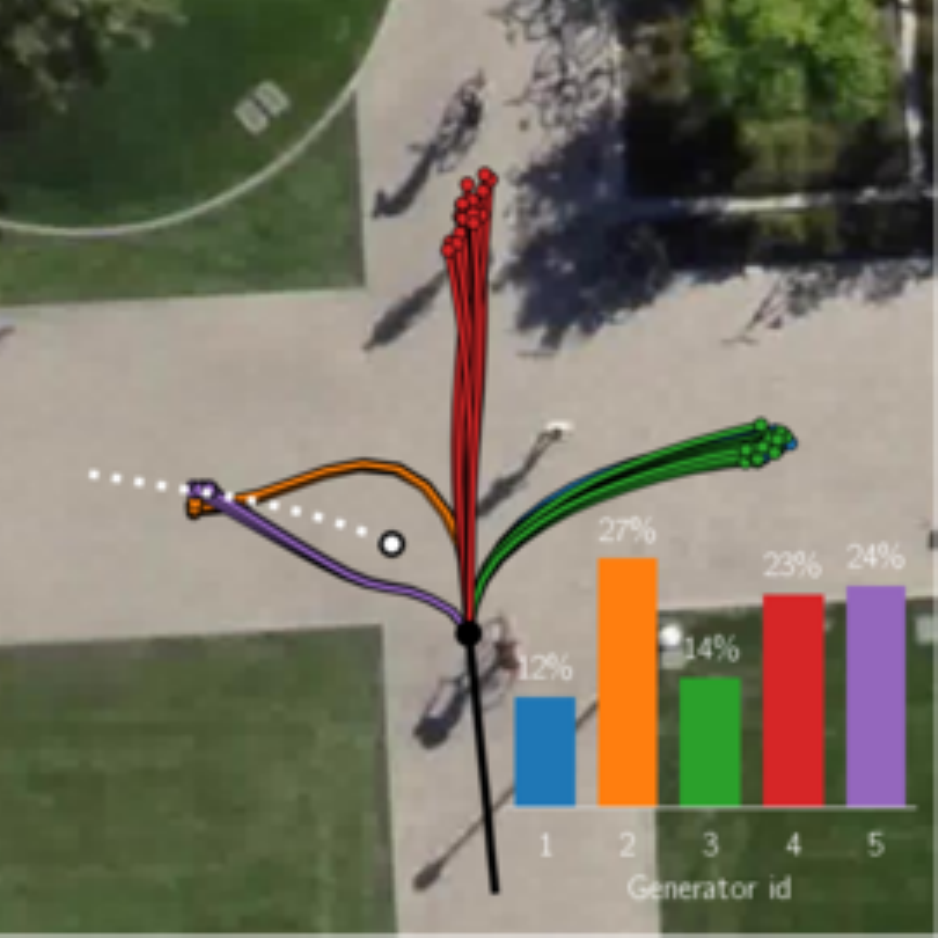}
        \includegraphics[width=\textwidth]{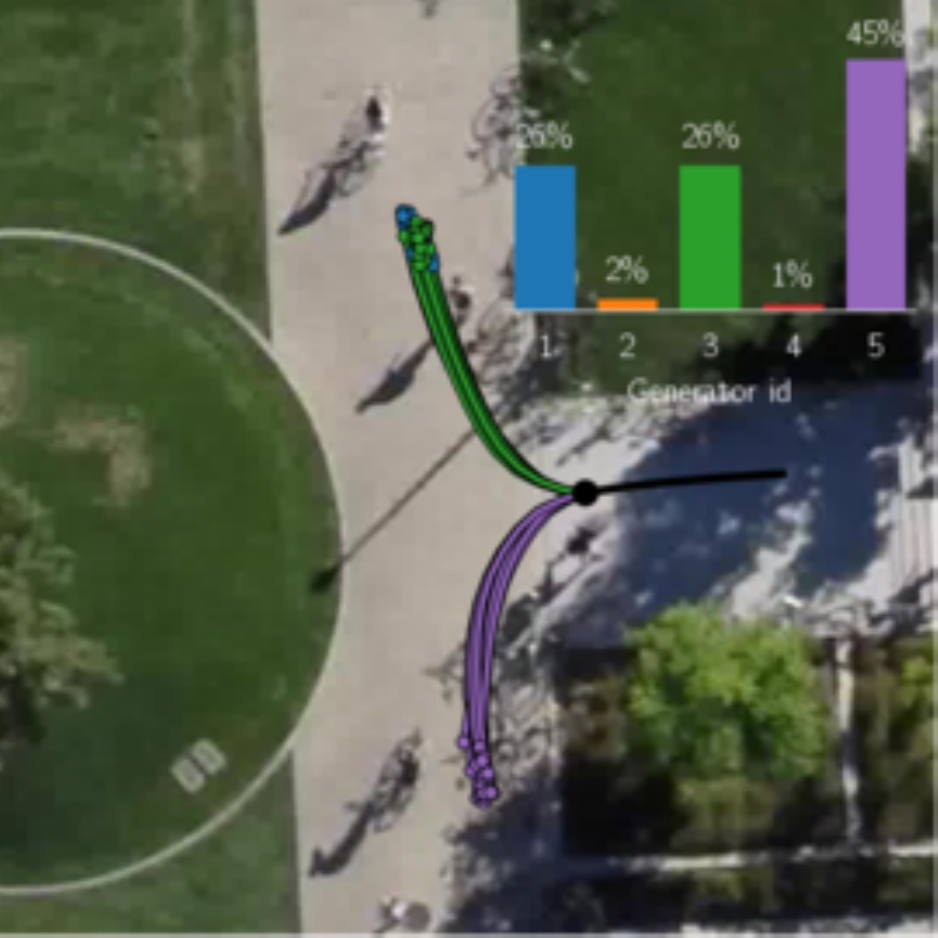}
        \caption{\modelname{} (ours)}
        \label{fig:synthetic:upperMGGAN}
    \end{subfigure}%
    \hfill
    \caption{Predicted trajectories for two scenarios in the synthetic dataset. The upper row contains scene on a junction with 3 modes and an interacting pedestrian (white). The lower row shows a scenario with two modes. Figures (a) represent the support of the conditional multimodal ground-truth distributions for these scenes. Figures (e) of \modelname{} also show the probabilities $\boldsymbol{\pi}$ of the PM Network. We visualize trajectories of one generator/discrete latent variable in the same color.}
    \label{fig:exp_standford_synthetic}
\end{figure*}

\subsection{Path Mode Network (\probnetname{})}
\label{sec:method:pinet}
The Path Mode Network (\probnetname{}) parameterizes a distribution over the indices of the generators $p(g|c) =[\pi_1, \cdots, \pi_{n_G}]$ conditioned on the features $c$ and is modelled with a multi-layer perceptron $\text{MLP}(c)$. The outputs $\{\pi_g\}$ assign probabilities to each of the $n_G$ generators. During inference, we can sample different generators based on the predicted distribution. Note, that this provides a major advantage over existing methods~\cite{mgan, disconnectedManifold}, where the distribution is fixed and cannot adjust to different scenes. In comparison, our \probnetname{} is capable of selecting the relevant generators for a given scene while deactivating unsuitable ones.

\subsection{Model Training}
We now present a training algorithm that jointly optimizes the distribution over generators parameterized by \probnetname{} and the multi-generator GAN model. For this, we propose an alternating training scheme, inspired by expectation-maximization~\cite{greff2017neural, disconnectedManifold}.

\subsubsection{GAN Training}
We train our model using a conditional generator,  discriminator network $D$~\cite{gans} that distinguishes between real and fake trajectories and a classifier $C$~\cite{mgan} learning to identify which generator predicted a given trajectory. 
More details on these networks' architectures can be found in the supplementary.
\vspace{\xxx}
\paragraph{Adversarial Loss.}
We define each generator $G_g$ as $\hat{Y}_{g,z}=G_g(c, z)$ inducing an implicit distribution $p_{G_g}(\hat{Y}| c)$. All $n_G$ generators together describe a joint probability distribution $\sum_{g=1}^{n_G} \pi_g \, p_{G_g}(\hat{Y}| c)$, thus the established results~\cite{gans} for GANs hold. We use the original adversarial loss $\mathcal{L}_{Adv}$~\cite{gans}. 
The discriminator $D$ learns to distinguish between real samples $Y$ and samples $\hat{Y}$ generated by the model encouraging realism of the predictions. However, $D$ by itself does not prevent the generators from collapsing to the same mode. 
\vspace{\xxx}
\paragraph{Classification Loss.} 
To incentivize the generators to cover different, possibly distinct modes occupying different regions of the output space, we follow \cite{mgan} and introduce a classifier $C$ which aims to identify the generator index $g$ that generated a sample $\hat{Y}_{g,z}$. 
The cross-entropy loss $\mathcal{L}_{Cl}$ between the classifier output and the true generator label of the predicted trajectory encourages the generators to model non-overlapping distributions and drives the trajectories of different generators spatially apart. This behavior is regularized through the adversarial loss $\mathcal{L}_{Adv}$ that constrains the samples to be realistic and not diverge from the real distribution.
Overall, the training object reads as follows
\begin{align}
    \min_G \max_D \mathcal{L}_{Adv} + \lambda_{Traj} \mathcal{L}_{Traj} + \lambda_{Cl} L_{Cl}, \label{eq:method:optimization}
\end{align}
where we additionally apply a $L_2$ best-of-many loss~\cite{best-of-many-sampling, social_gan} $\mathcal{L}_{Traj}$ with $q$ samples to increase the diversity of predicted trajectories. $\lambda_{Traj}$ and $\lambda_{Cl}$ are weighting hyperparameters.

\subsubsection{\probnetname{} Training}
\label{sec:method:subsec:optmization:subsubsec:pinet_training}

To train \probnetname{}, we approximate the likelihood of a particular generator distribution $p_{G_g}$ supporting the trajectory $Y$ by the generated trajectories $\hat{Y}_{g, c, z_i} = G_g(c, z_i)$ as: 
\begin{equation}
    p(Y|c, g) \propto \frac{1}{l}\sum_{i = 1}^l \exp \left( \frac{- \norm{\hat{Y}_{g, c, z_i}- Y}_2^2}{2\sigma} \right). 
\end{equation}
Here, we marginalize the GAN noise $z$ and assume a normally distributed and additive error $\epsilon \sim N(0, \sigma I)$ between $\hat{Y}$ and $Y$ as common for regression tasks~\cite{regressionanalysis}. 
We obtain the conditional probability over generators by applying Bayes' rule:
\begin{equation}
     p(g |c, Y) = \frac{ p(Y|c, g)} {\sum_{g^\prime}^{n_G}  p(Y|c,{g^\prime})}  \label{equ:method:endpoint_approximation}.
\end{equation}

Finally, we optimize the \probnetname{} with the approximated likelihood minimizing the cross entropy loss:
\begin{equation}
    \mathcal{L}_{\Pi} = H(p(g|c, Y),\Pi(c)). \label{eq:method:pinet_loss_supp}
\end{equation}
Intuitively, the network is trained to weigh the generator that generates trajectories closest to the ground-truth sample the highest.
We provide the full derivation of the objective in the supplementary.
\subsubsection{Alternating Training Scheme}
Our training scheme consists of two alternating steps similar to an expectation-maximization algorithm~\cite{greff2017neural}:
\vspace{\xxx}
\paragraph{1. \probnetname{} Training Step:} We sample $l$ trajectories for each generator and optimize the parameters of \probnetname{} using \Cref{eq:method:pinet_loss_supp} while keeping the rest of the network's parameters fixed. 
\vspace{\xxx}
\paragraph{2. Generator Training Step:} In the generator training step, we use \probnetname{} to generate probabilities $\boldsymbol{\pi}$ and sample $q$ generators predicting trajectories. With these predictions, we update the model excluding \probnetname{} optimizing \Cref{eq:method:optimization}. We provide pseudo-code detailing our training procedure in the supplementary. 

\subsection{Trajectory Sampling} 
\label{sec:method:sampling}
We can use the estimated probabilities $\boldsymbol{\pi} = [\pi_1, \dots, \pi_{n_G}]$ generated by the \probnetname{} to establish different mechanisms to sample trajectories from the multiple generators. This helps us to cover all modes present in the scene with as-few-as-possible predictions.
In single generator models~\cite{bigat, sadeghian2018sophie} the relation between regions in the Gaussian latent space and different modes in the output space is implicit and unknown. 
However, for \modelname{} we can use the estimated probabilities $\boldsymbol{\pi} = [\pi_1, \dots, \pi_{n_G}]$ from the \probnetname{} to control and to cover predictions for all modes present in the scene.
Next to randomly sampling  $k$ trajectories (\textit{Random}) from $\boldsymbol{\pi}$ we introduce an additional strategy (\textit{Expectation}) where we compute the expected number of samples for each generator as $n_g = k \cdot \pi_g$. We round all $n_g$ to the nearest integer and adjust the number of the generator with the highest score to ensure that all numbers sum up to $k$.

\begin{figure}
\centering
\includegraphics[width=\linewidth]{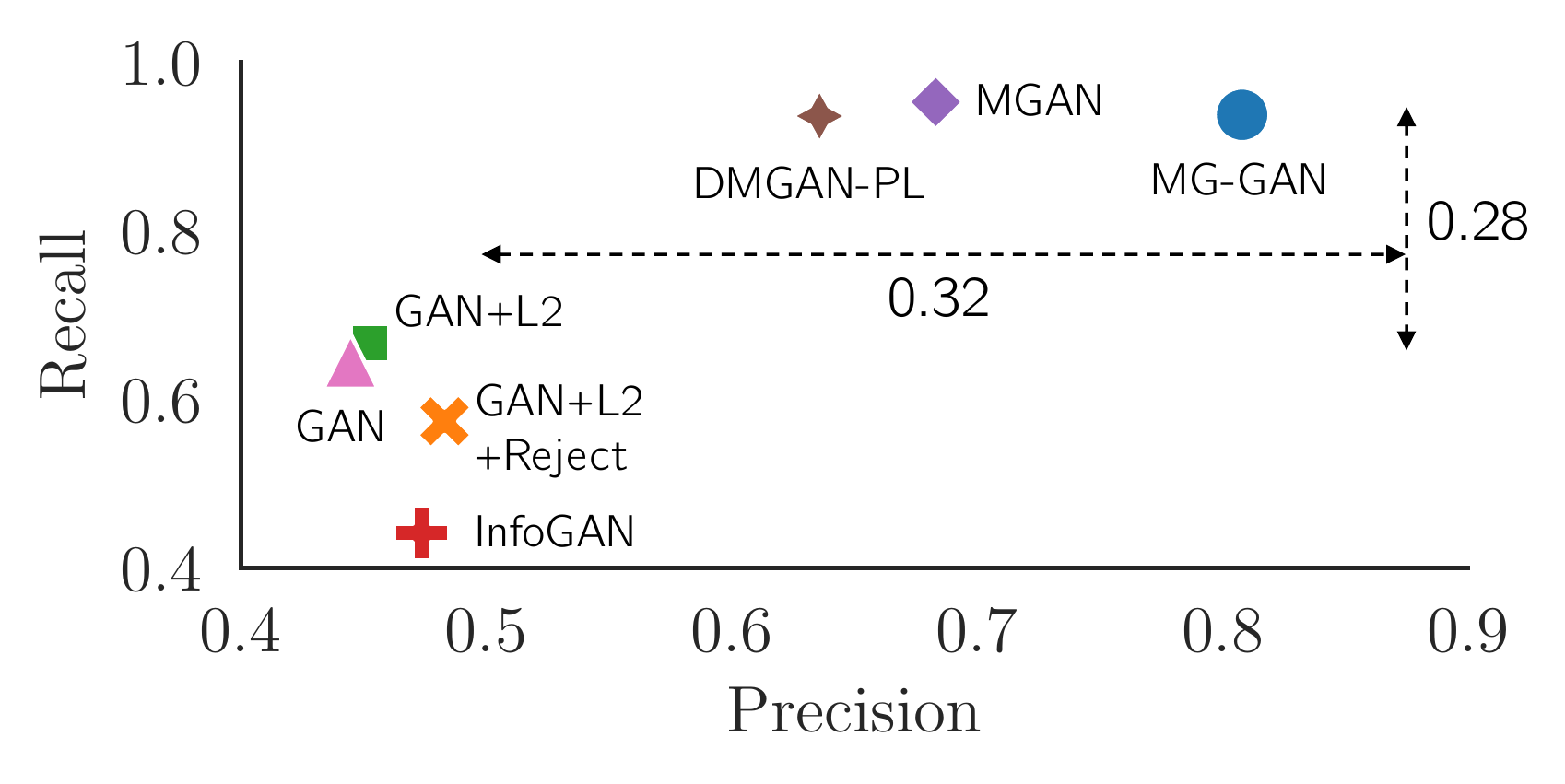}
\caption{Precision vs. Recall on synthetic dataset.}\label{fig:synthetic:recallprecision}
\end{figure}

\section{Experimental Evaluation}
\label{sec:experiments}
We evaluate our model on four publicly available datasets~\cite{ETH-data, UCY-data, stanforddronedataset,liang2020garden} for pedestrian trajectory prediction and compare our results with state-of-the-art methods. Furthermore, we conduct experiments on synthetic datasets. Compared to real data, synthetic data provides access to the ground-truth trajectory distribution which enables us to identify OOD samples by comparing ground-truth and generated trajectory distributions. Finally, we run an ablation on the individual components of \modelname{} and study the robustness of our model \wrt the number generators $n_G$.

\subsection{Experimental Setup}
\label{sec:sub:setup}

We follow prior work~\cite{stanforddronedataset, sociallstm} and observe 8 past time steps (3.2 seconds) and predict the future 12 time steps (4.8 seconds) for every pedestrian in the scene.
\vspace{\xxx}
\paragraph{Metrics.} We evaluate results using the following metrics: \newline
\noindent{\it Average Displacement Error ({ADE})} is defined as a mean $L_2$ distance between the prediction and ground-truth trajectory.\newline
\noindent{\it Final Displacement Error ({FDE})} is defined as the distance between the prediction and ground-truth trajectory position at time $t_{pred}$. \newline
For both metrics, ADE and FDE, we follow the \textit{Minimum over k} procedure~\cite{social_gan, sadeghian2018sophie, bigat} with $k=20$. 
Note that this approach only considers a single prediction with the lowest ADE and FDE, but not the entirety of the set of $k$ generated output trajectories combined. Therefore, we include additional metrics commonly used in the GAN literature~\cite{FirstPrecisionRecallNIPS, ImprovedPrecisionRecallNIPS}, namely \textit{recall} and \textit{precision}. Recall measures the coverage of all ground-truth modes, while precision measures the ratio of generated samples in the support of the ground truth distribution. Hence, the precision is directly related to the number of OOD samples. We also compute the $F1$ score, combining recall and precision.
\vspace{\xxx}
\paragraph{Datasets.} 
We perform the evaluation using the following datasets. ETH~\cite{ETH-data} and UCY datasets~\cite{UCY-data} contain five sequences (ETH: ETH and HOTEL, UCY: UNIV, ZARA1, and ZARA2), recorded in four different scenarios. We follow the standard leave-one-out approach for training and testing, where we train on four datasets and test on the remaining one. 
The Stanford Drone Dataset (SDD)~\cite{stanforddronedataset} consists of $20$ video sequences captured from a top view at the Stanford University campus. In our experiments, we follow the train-test-split of~\cite{trajnet} and focus solely on pedestrians. 
The recently proposed Forking Paths Dataset (FPD)~\cite{liang2020garden} is a realistic 3D simulated dataset providing multi-future trajectories for a single input trajectory. 
To study the ability of our model to predict multimodal trajectories while preventing OOD samples, we create a synthetic dataset where we simulate multiple possible future paths for the same observation emerging due to the scene layout and social interactions. Detailed information on the generated dataset is provided in the supplementary material. 
\vspace{\xxx}
\paragraph{Baselines.}
We compare our method with several single and multi-generator GAN baselines. We evaluate a (i) vanilla \textit{GAN} baseline, (ii) \textit{GAN L2} trained with variety loss~\cite{best-of-many-sampling}, (iii) \textit{GAN L2 Reject}~\cite{noGANlandICML} that filters OOD samples based on gradients in the latent space, and (iv) InfoGAN~\cite{infogan} with discrete random latent variable.
Furthermore, we compare \modelname{} to multi-generator models MGAN~\cite{mgan} and DMGAN-PL~\cite{disconnectedManifold}, proposed in the context of image generation, that we adapt for the task of trajectory prediction. To ensure comparability, all models use the same base model following SoPhie \cite{sadeghian2018sophie} with attention modules as described in \Cref{sec:method:mgan}.
For qualitative comparison, we evaluate our method against state-of-the-art prediction models presented in \Cref{sec:related_work} on the standard benchmarks for trajectory forecasting. 

\subsection{Experiments on Synthetic data}
\label{sec:experiments:subsec:synthetic}
\begin{figure}
    \begin{subfigure}[b]{0.32\linewidth}
    \includegraphics[width=\linewidth]{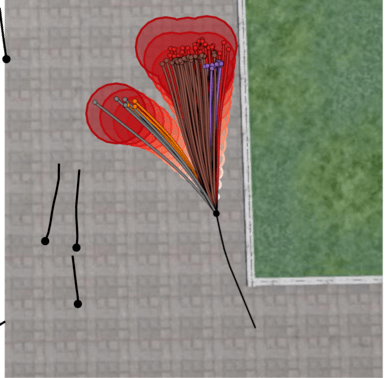}
    \end{subfigure} 
    \begin{subfigure}[b]{0.32\linewidth}
    \includegraphics[width=\linewidth]{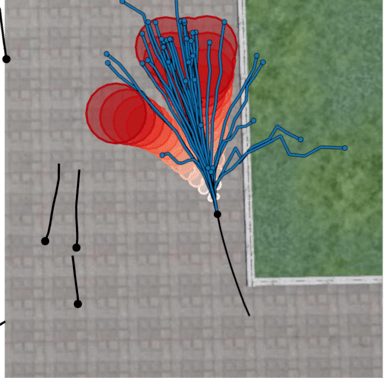}%
    \end{subfigure}
    \begin{subfigure}[b]{0.32\linewidth}
    \includegraphics[width=\linewidth]{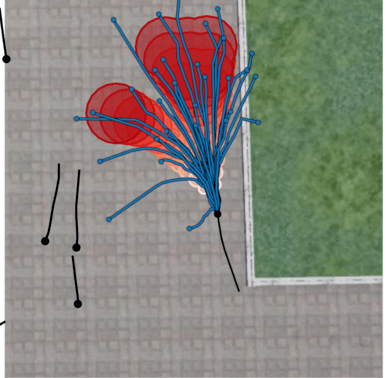}
    \end{subfigure}
    \hfill
    \caption{Generated samples of our \modelname{}, Trajectron++, and PECNet.}
    \label{fig:gofp_samples_sota}
\end{figure}
We first study our model on a synthetic dataset in which we have access to the ground-truth distribution of the future trajectories. In this experiment, we show that \modelname{} achieves better performance in learning a multimodal trajectory distribution with disconnected support and is more efficient than the baselines.
\vspace{\xxx}
\paragraph{Results.} 
The results in \Cref{fig:synthetic:recallprecision} show that \modelname{ } outperforms the single-generator baselines and increases Recall by $0.28$ and Precision by $0.32$.
To this end, we find that all multi-generator methods have a similar recall but \modelname{} achieves a $15\% $ higher Precision corresponding to a lower number of OOD samples.
\begin{figure*}
     \centering
\subfloat
  []
  {\label{fig:synthetic:samplingk} \includegraphics[width=0.27\textwidth]{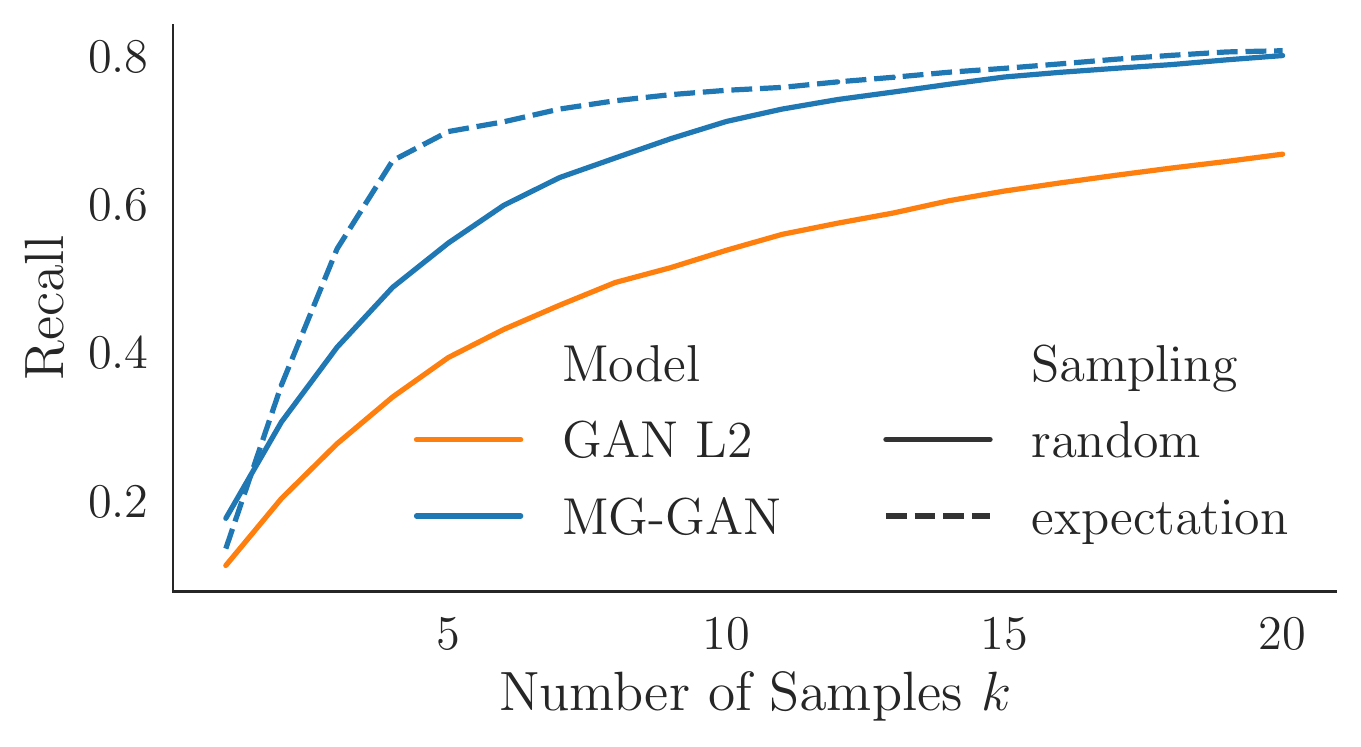}}
\hfill
\subfloat
  []
 {\label{fig:synthetic:num_weights:ADEFDE}
 \includegraphics[width=0.23\textwidth]{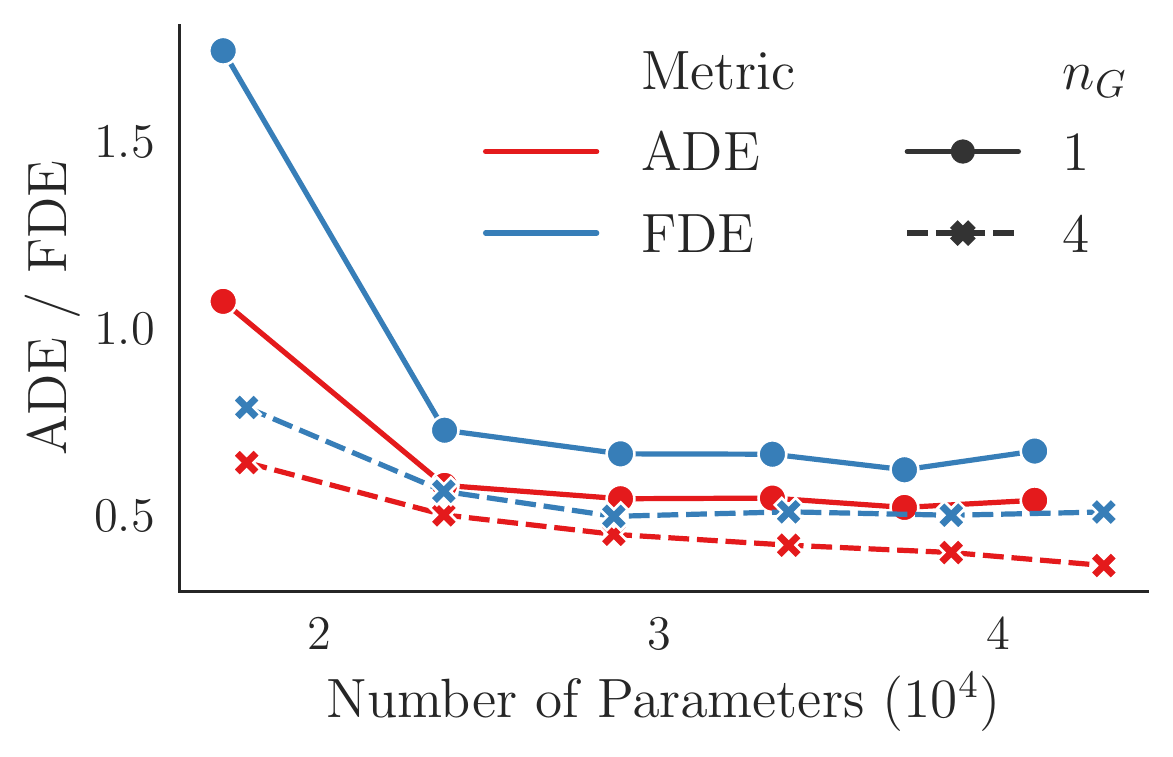}}
 \hfill
\subfloat
  []
 {\label{fig:synthetic:num_weights:RECALLPRECISSION} \includegraphics[width=0.23\textwidth]{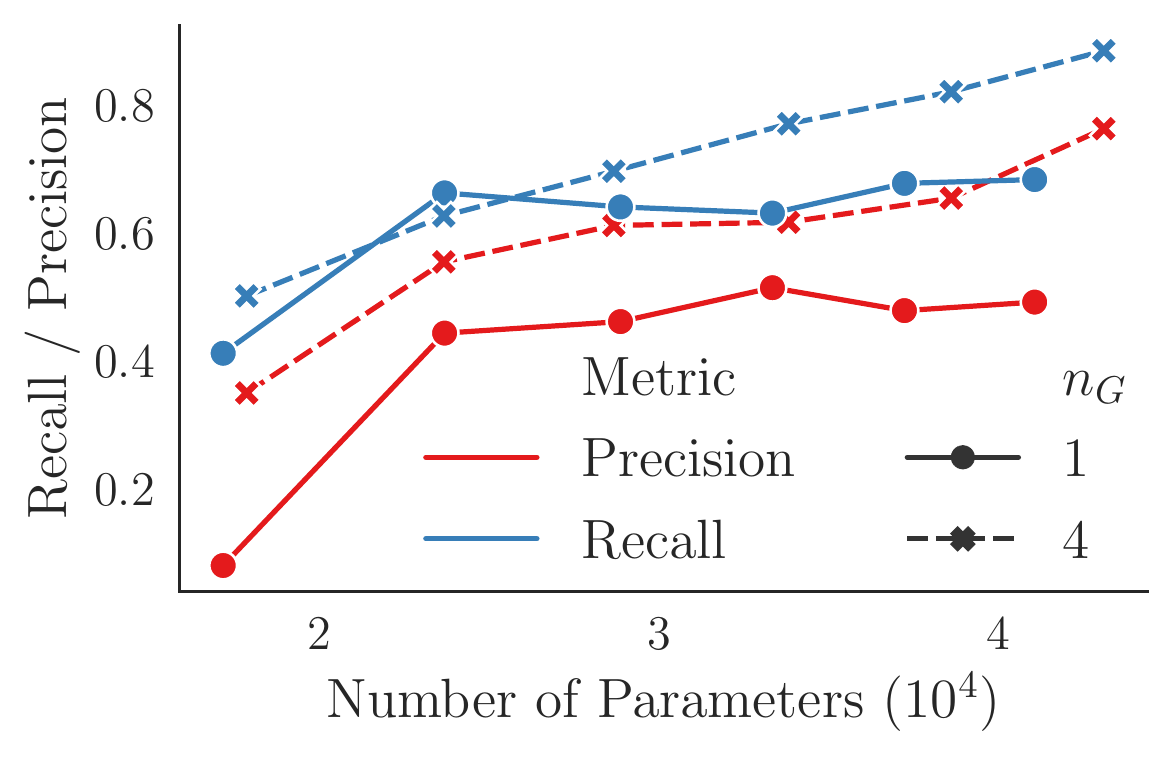}}
 \hfill
\subfloat
  []
 {\label{fig:synthetic:num_weights:MACS} \includegraphics[width=0.23\textwidth]{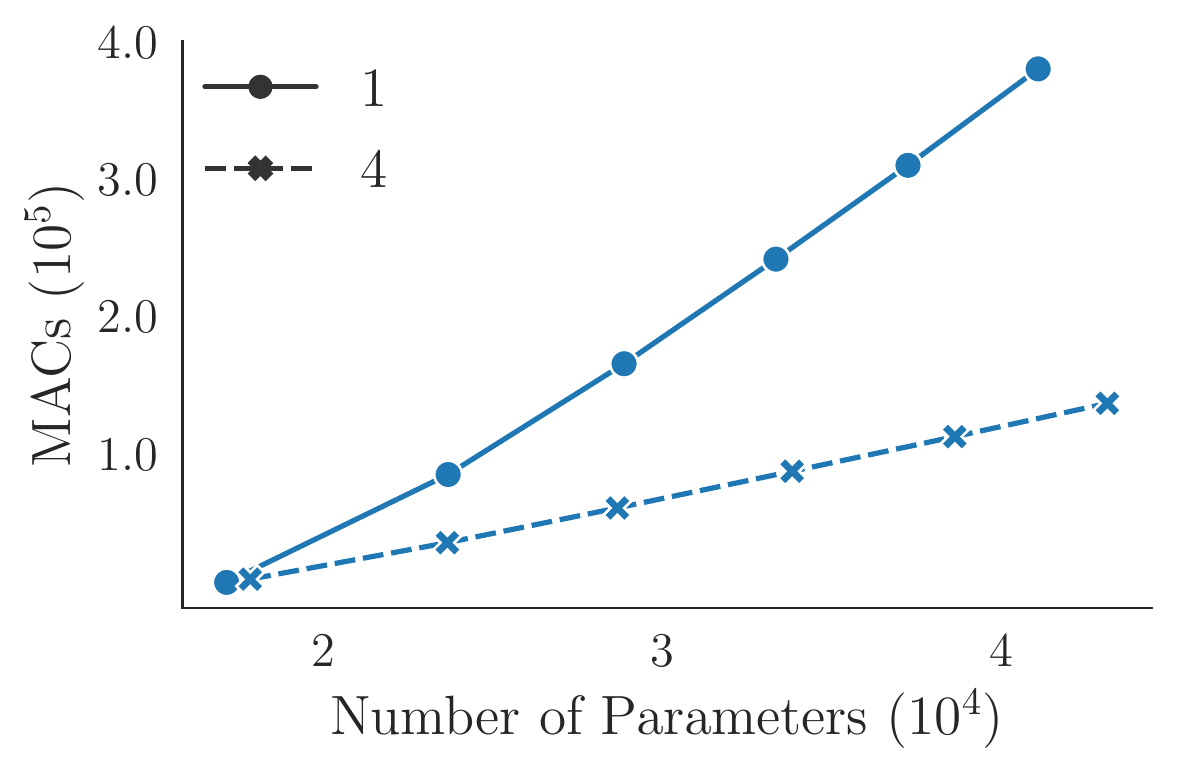}}
\caption{Comparison between single generator model \textit{GAN+L2} and \modelname{}. (a) recall for different number of samples $k$ and sampling methods. (b) - (c) compares ADE/FDE, recall/precision, and MACs (Multiply–accumulate operations) for varying total number of model parameters.}
\end{figure*}

\vspace{\xxx}
\paragraph{ Visual Results.} 
In \Cref{fig:exp_standford_synthetic}, we visualize predicted trajectories for two different scenarios where the white trajectory represents another interacting pedestrian. The support of the ground-truth distribution for each timestep is shown as a red circle in \Cref{fig:synthetic:upperGT}. A model achieves low precision in \Cref{fig:synthetic:recallprecision} if many trajectory points lie outside the corresponding red circle for a particular timestep. Similarly, a model has high recall if its samples cover most of the area of red circles.
Single generator models, GAN+L2 (\Cref{fig:synthetic:upperGANL2}), and InfoGAN (\Cref{fig:synthetic:upperInfoGAN}) produce many OOD samples leading to low precision. In particular, we find that InfoGAN is not able to learn the correspondence between the discrete latent space and the modes in the trajectory space. While theoretically plausible, these results indicate that a discretized latent space is not well-suited for learning distribution on disconnected support.
Contrarily, MGAN can learn the distribution but is incapable to adjust generators resulting in OOD samples in \Cref{fig:synthetic:upperMGAN} when the number of modes does not match the number of generators. 
Finally, our \modelname{} is able to adjust to both scenarios in \Cref{fig:synthetic:upperMGGAN} as the \probnetname{} deactivates generators which are unsuitable and prevents OOD samples explaining the high Precision in \Cref{fig:synthetic:recallprecision}.
\vspace{\xxx}
\paragraph{ Effective Mode Covering.} \Cref{fig:synthetic:samplingk} shows the recall depending on the number of samples $k$. 
Our method covers more modes of the ground-truth distribution than the single generator model for the same number of samples as indicated by the higher recall. Additionally, we observe significant improvements compared to random sampling by using expectation sampling leveraging \probnetname{} as described in \Cref{sec:method:sampling}, especially for fewer samples.
\vspace{\xxx}
\paragraph{ Number of Parameters and Computational Cost.}
In this experiment, we show that our \modelname{} does not require more resources \wrt parameters or computations compared to a single generator baseline.
For this, we compare \modelname{} using four generators with the single-generator baseline while keeping the total number of parameters of both models fixed by only using approx. $\nicefrac{1}{4}$ of the parameters for each generator.
As can be seen in \Cref{fig:synthetic:num_weights:ADEFDE,fig:synthetic:num_weights:RECALLPRECISSION}, \modelname{} outperforms the single generator GAN \wrt to ADE/FDE ($50\%)$ and recall/precision ($30\%)$ using the same number of total parameters across various parameter budgets.
In \Cref{fig:synthetic:samplingk}, the computational cost measured by MACs for the prediction of a trajectory is always lower for \modelname{} compared to the baseline. The model only runs one selected generator with $\nicefrac{1}{4}$  amount of parameters during the forward pass while the cost of running \probnetname{} is negligible. 
\subsection{Benchmark Results}
\label{sec:experiments:subsec:benchmark}

In this section, we compare our method to the state-of-the-art on the standard benchmarks ETH~\cite{ETH-data}, UCY~\cite{UCY-data}, and SDD~\cite{stanforddronedataset}, as well as the recently proposed Forking Path Dataset (FPD)~\cite{liang2020garden}. 
We report the performance of the model with the lowest validation error as we train our method with different numbers of generators $n_G \in \{2, \dots, 8\}$. We discuss the robustness \wrt the number of generators in \Cref{sec:experiments:subsec:ablation_study}.

\begin{table}
\centering
\setlength{\tabcolsep}{1mm}
\resizebox{\linewidth}{!}{%
    \begin{tabular}{l*{10}{c}}\toprule
Dataset&
\makecell{ S-LSTM\\\cite{sociallstm}}&
\makecell{ S-GAN\\\cite{social_gan}}&
\makecell{ SoPhie\\\cite{sadeghian2018sophie}}&
\makecell{ S-BiGAT\\\cite{bigat}}& 
\makecell{ CGNS\\\cite{li2019conditional}}&
\makecell{ GoalGAN\\\cite{dendorfer20accv}}& 
\makecell{ PECNet\\\cite{mangalam2020pecnet}}& 
\makecell{ Trajectron++\\\cite{Salzmann2020TrajectronDT}}& 
\makecell{ MG-GAN\\ (Ours) } \\
\midrule
\textbf{ETH}    &  1.09/2.35 &  0.81/1.52 &  0.70/1.43 &  0.69/1.29 &  0.62/1.40 &  0.59/1.18 &              0.54/\underline{0.87} &  \bfseries{0.39}/\bfseries{0.83} &              \underline{0.47}/0.91 \\
\textbf{HOTEL}  &  0.79/1.76 &  0.72/1.61  &  0.76/1.67 &  0.49/1.01 &  0.70/0.93 &  0.19/0.35 &              0.18/\underline{0.24} &  \bfseries{0.12}/\bfseries{0.21} &  \underline{0.14}/\underline{0.24} \\
\textbf{UNIV}   &  0.67/1.40 &  0.60/1.26  &  0.54/1.24 &  0.55/1.32 &  0.48/1.22 &  0.60/1.19 &  \underline{0.35}/\underline{0.60} &  \bfseries{0.20}/\bfseries{0.44} &                          0.54/1.07 \\
\textbf{ZARA1}  &  0.47/1.00 &  0.34/0.69  &  0.30/0.63 &  0.30/0.62 &  0.32/0.59 &  0.43/0.87 &  \underline{0.22}/\underline{0.39} &  \bfseries{0.15}/\bfseries{0.33} &                          0.36/0.73 \\
\textbf{ZARA2}  &  0.56/1.17 &  0.42/0.84  &  0.38/0.78 &  0.36/0.75 &  0.35/0.71 &  0.32/0.65 &  \underline{0.17}/\underline{0.30} &  \bfseries{0.11}/\bfseries{0.25} &                          0.29/0.60 \\
\midrule
\textbf{AVG}    &  0.72/1.54 &  0.58/1.18  &  0.54/1.15 &  0.48/1.00 &  0.49/0.97 &  0.43/0.85 &  \underline{0.29}/\underline{0.48} &  \bfseries{0.19}/\bfseries{0.41} &                          0.36/0.71 \\
\bottomrule
\end{tabular}

}

\caption{Quantitative results on ETH~\cite{ETH-data} and UCY~\cite{UCY-data}. We report ADE ($\downarrow$) /FDE ($\downarrow$) in meters. Underlined results denote the second best.}
\label{table:experiments:biwi}
\end{table}

\begin{table}
\centering

\setlength{\tabcolsep}{1mm}
\resizebox{\linewidth}{!}{%
    \begin{tabular}{lccccccccccc} \toprule
&
\makecell{S-LSTM\\\cite{sociallstm}}&
\makecell{S-GAN\\\cite{social_gan}}&
\makecell{CAR-NET\\\cite{sadeghian2017carnet}}&
\makecell{DESIRE\\\cite{desire}}&
\makecell{SoPhie\\\cite{sadeghian2018sophie}}&
\makecell{CGNS\\\cite{li2019conditional}}&
\makecell{CF-VAE\\\cite{bhattacharyya2019conditional}}&
\makecell{P2TIRL\\\cite{deo2020trajectory}}&
 \makecell{GoalGAN\\\cite{dendorfer20accv}} &
 \makecell{PECNet\\\cite{mangalam2020pecnet}} &
 \makecell{MG-GAN (4)\\ (Ours) } \\
\midrule
\textbf{ADE} &57.0 &27.3&25.7 &19.3&16.3 & 15.6 & 12.6 & 12.6 &  12.2   & \textbf{10.0}          &  13.6 \\
\textbf{FDE} &31.2 &41.4 &51.8 &34.1 &29.4 & 28.2 & 22.3 & 22.1 & 22.1 & \textbf{15.9} &  25.8 \\
\bottomrule \end{tabular}
}
\caption{Quantitative results on Stanford Drone Dataset (SDD)~\cite{stanforddronedataset}. We report ADE and FDE in pixels. }\label{Table:SDD}
\end{table}

\vspace{\xxx}
\paragraph{ADE \& FDE.}
Our \modelname{} achieves competitive results for the ADE and FDE on the ETH/UCY and Stanford Drone Dataset (SDD) shown in \Cref{table:experiments:biwi} and \Cref{Table:SDD}, respectively.
Even though our method does not achieve SOTA performance on the ADE and FDE metrics on these benchmarks, we still argue that our method provides significant improvement to the task. That is since the distance-based 
\(L_2\) measures can be drastically reduced by increasing the variance of the predictions for the price of producing more OOD samples. A visual comparison of the trajectories produced by Trajectron++ and PECNet in \Cref{fig:gofp_samples_sota} shows that these methods produce high variance predictions without accounting for any constraints in the scene. Contrarily, \modelname{}  only predicts trajectories inside the ground-truth manifold (red). While covering all modes, our predictions remain in the support of the ground-truth distribution.
To quantify this observation, we compute the recall and precision metrics.
\vspace{\xxx}
\paragraph{Precision \& Recall.}
\label{sec:experiment:recallprecision}
As ADE and FDE do not consider the quality of the entire generated distribution, we add results using precision/recall metrics~\cite{FirstPrecisionRecallNIPS, ImprovedPrecisionRecallNIPS} on the FPD dataset~\cite{liang2020garden}. This is possible on FPD as it contains multiple feasible, human-annotated ground-truth trajectories.
\begin{table}
\centering
\setlength{\tabcolsep}{1mm}
\resizebox{\linewidth}{!}{%
\begin{tabular}{lrrccc} 
\toprule
              & ADE $\downarrow$ & FDE $\downarrow$ & Precision $\uparrow$ & Recall $\uparrow$ & F1 $\uparrow \quad$  \\ 
\midrule
GAN+L2        & 28.81                  & 58.37                  & 0.55                 & 0.87                  & 0.67                 \\
\midrule
PECNet        & \textbf{13.14}         & \textbf{24.55}         & 0.46                 & 0.95                  & 0.62                 \\
Trajectron++  & 13.15                  & 32.00                  & 0.38                 & \textbf{0.96}         & 0.54                 \\ 
\midrule
MG-GAN (Ours) & 22.09                  & 46.38                  & \textbf{ 0.71}       & 0.89                  & \textbf{0.79}        \\
\bottomrule
\end{tabular}
}
\caption{Results on FPD. We report ADE/FDE in pixels.}\label{Table:precision_recall}
\end{table}

In \Cref{Table:precision_recall}, \modelname{} outperforms \textit{GAN+L2} by $29\%$, PECNet by \(54\%\) and Trajectron++ by \(86\%\) in terms of Precision, while the difference in Recall with $0.02$, \(0.06\), and \(0.07\) points is small. Single generator models predict overly diverse trajectories, thus increasing Recall slightly and reducing ADE/FDE, but produce OOD samples leading to low Precision. 
These results confirm that \modelname{} is significantly more reliably at predicting paths that align well with the human-annotated future trajectories (high precision), while also covering a similar amount of modes in the scene (high recall). 
Overall, we conclude that \modelname{} does not match SOTA performance on traditional evaluation metrics in \Cref{table:experiments:biwi} and \Cref{Table:SDD}. However, studying precision and recall reveals that our model can lower the number of OOD and achieves an overall better \textit{F1} than current SOTA methods.

%
\subsection{Ablation Studies}
\label{sec:experiments:subsec:ablation_study}
In this section, we ablate the key modules of \modelname{}.
We emphasize that the goal of the paper is to demonstrate the need and effectiveness of a conditional multi-generator framework for pedestrian trajectory prediction. Hence, the study of attention modules used within our model described in \Cref{sec:method:mgan}, is not the goal of this work and has been extensively done in prior work~\cite{social_gan,social_ways, sadeghian2018sophie, bigat}. 

\begin{table}[!htbp]
\centering
\small
\setlength{\tabcolsep}{1mm}
\begin{tabular}{lllrrrr} 
\toprule
 M & C & PM              & ADE $\downarrow$ & FDE $\downarrow$ & Precision $\uparrow$ & Recall $\uparrow$  \\ 
\midrule
 & & &  0.94 	& 1.58 	&    0.46   &   0.48      \\
\checkmark  & &   & 0.59   &  0.79 &    0.37 & 0.68 \\
\checkmark & & \checkmark  &  0.35   &  0.49                         &    0.72 & 0.91 \\
\checkmark & \checkmark &  & 0.37   &  0.53 &    0.73 &  0.91     \\
\midrule
\checkmark & \checkmark&\checkmark     &  \textbf{0.32}   &  \textbf{0.44} &  \textbf{ 0.77} &   \textbf{0.95}      \\
\bottomrule
\end{tabular}
\caption{Ablation experiments: (M) Multi-generator, (C) Classifier, and (PM) Path Mode network. }
\label{table:experiments:ablation}
\end{table}
\vspace{\xxx}
\paragraph{Effectiveness of Key Modules.}
We perform the ablation on our synthetic dataset by removing key components from our final model: multiple generators, the classifier $C$, and the \probnetname{} in \Cref{table:experiments:ablation}.
Reducing the number of generators to \(1\) results in a significant drop in performance of almost $ 50\%$ in recall and $31\%$ in precision.

As described in \Cref{sec:method:mgan}, the classifier $C$ encourages individual generators to specialize and increases precision from $37\%$ to $73\%$. Similarly, with \probnetname{} learning a distribution over generators, the precision increases from $37\%$ to $72\%$.
Finally, leveraging \probnetname{} and classifier \(C\), combining the advantages of both, further improves the performance on all considered metrics.
\vspace{\xxx}
\paragraph{Robustness over the Number of Generators.}
The multimodality over future trajectories depends on social interactions and the scene layout, imposing a significant challenge when choosing the number of generators \(n_G\) at training time. To this end, we introduced the \probnetname{} that learns to activate generators depending on the observed scene features. 
As can be seen in \Cref{Table:results_num_gnerators}, \probnetname{} successfully makes \modelname{} robust \wrt the choice of \(n_G\) as results only deviate $7\%$ from the best reported values at maximum. 
\begin{table}
\centering
\small
\setlength{\tabcolsep}{1mm}
\begin{tabular}{lrrrrrrrr}
\toprule
 &          2 &          3 &          4 &          5 &  6 & 7 & 8 &     Best \\
\midrule
ADE      &  0.37 &  0.38 &  0.38 &  0.39  & 0.37 & 0.36 & 0.37 & 0.36\\
FDE          & 0.72 &  0.74 &  0.75 & 0.76& 0.71 & 0.71 & 0.72  & 0.70   \\
\bottomrule
\end{tabular}
\caption{Results for $n_G \in \{2, \dots, 8\}$ on ETH/UCY.}
\label{Table:results_num_gnerators}
\end{table}
\vspace{\xxx}
\section{Conclusion}
In this paper, we addressed the issue of single-generator GAN models for pedestrian trajectory prediction. While existing generative networks learn a distribution over future trajectories, they are fundamentally incapable of learning a distribution consisting of multiple disconnected modes. To overcome this problem, our proposed \modelname{} leverages multiple generators that specialize in different modes and learns to sample from these generators conditioned on the scene observation. We demonstrated the efficacy of \modelname{} at reducing out-of-distribution samples in comparison to the existing state-of-the-art. Finally, we emphasized the importance of precision next to recall metrics and hope to encourage a discussion on preventing OOD in future work.
\newline \textbf{Acknowledgements.} This project was funded by the Humboldt Foundation through the Sofja Kovalevskaja Award. We highly thank Aljo\v{s}a O\v{s}ep for helpful discussions, constructive feedback, and proofreading.

{\small
\bibliographystyle{ieee_fullname}
\bibliography{egbib}
}


\clearpage
\appendix
\renewcommand{\thesection}{\Alph{section}.\arabic{section}}
\setcounter{section}{0}

\begin{appendices}
The supplementary material complements our work with the implementation details of \modelname{} in \Cref{sup:sec:architecture}. Furthermore, we provide details on the training procedure and additional experiments on the hyperparameters of \modelname{} in \Cref{sup:sec:training}. In \Cref{sec:precisionrecall}, we explain how we compute Precision and Recall and describe the synthetic dataset in \Cref{supp:section:synthetic_datset}. We discuss the performance of \modelname{} on real-world datasets in \Cref{sup:sec:multimodality_real_datasets}. Lastly, we investigate the benefits of multi-generator models in learning distinct modes on a toy dataset in \Cref{sup:sec:toy_experiment} and add visualizations of predicted trajectories on the considered datasets in \Cref{sup:sec:visualizations}.

\section{Architecture}
\label{sup:sec:architecture}
The main contribution of our method is the use of multiple generators and the proposed Path Mode Network (\probnetname{}). The architecture of the individual generators is a standard Long Short-Term Memory (LSTM) encoder-decoder model~\cite{sociallstm} with social and physical attention as proposed in~\cite{sadeghian2018sophie,social_ways}. 
The entire model is trained in a GAN framework and uses a discriminator and additional classifier that encourages the generators to specialize to a specific mode. 

In the following paragraphs, we describe the architecture of all our components to generate a set of $K$ future trajectories $\{\hat{Y}^k_i\}_{k=1, \dots, K}$ with $t \in \left[t_{obs}+ 1, t_{pred} \right]$ given the input trajectory $X_i$ with $t \in \left[t_{1}, t_{obs} \right]$ for each pedestrian $i$. The source code of our model is provided with the supplementary material.
 
\subsection{Encoding}
\label{sup:sec:architecture:encoding}
To extract dynamic features from the past trajectory of a pedestrian $i$ in a scene, we use an LSTM~\cite{Hochreiter:1997:LSM:1246443.1246450} to encode the relative displacements $\Delta X_i$ into a high dimensional feature representation $d_i$
\begin{equation}
d_i = \LSTM_{en}\left(\Delta X_i^t, h_i^t\right), \nonumber
\end{equation}
where $h_i^t$ is the hidden state of the LSTM. We compute visual features $f_i$ with a CNN for an image patch $I_i$ cut around the last observed position of pedestrian $i$

\begin{equation}
 f_i = \text{CNN}\left(I_i\right). \nonumber
\end{equation}

The image patch $I_i$ is a $32 \times 32$ pixel patch with a resolution of $0.7 \nicefrac{m}{pixel}$. The CNN has $2$ layers with $16$ filters, kernel size $3$, max-pooling and ReLU activations and is trained from scratch.

The scene layout as well as other interacting pedestrians affect the path of a pedestrians and have to be considered by the model. For our method, we model physical (agent-scene) and social (agent-agent) interaction with corresponding soft-attention modules~\cite{ShowAttendTell}, following~\cite{sadeghian2018sophie, social_ways}.

\paragraph{Social Attention~\cite{social_ways}: }
To account for social interaction, we apply soft-attention on the hidden states $\{d_j\}_{j \in J}$ of the other pedestrians in the scene where we compute the attention score $a_{ij}$ based on the distance and the bearing angle between the agent $i$ and a neighbouring agent $j$. The social information for pedestrian $i$ is then defined as
\begin{equation}
 s_i = \sum_{j \in J} a_{ij} d_j
\end{equation}

\paragraph{Physical Attention~\cite{sadeghian2018sophie}: }
The interaction with the scene around a pedestrian is also modeled with a soft-attention network~\cite{ShowAttendTell} $ATT$ applied on the CNN features $f_i$ based on the motion encoding $d_i$. Thus, the physical features $v_i$ are given by

\begin{equation}
 v_i = \ATT\left( f_i, d_i \right)
\end{equation}

\paragraph{Final Encoding.}
Finally, the dynamic features $d_i$, social features $s_i$, and physical features $v_i$ are concatenated to form the conditional encoding $c_i$ for pedestrian $i$ which is used for generating predictions in the following. All vectors $d_i$, $v_i$, and $s_i$ have length $32$.

\subsection{Generator}
For \modelname{}, we propose $n_G$ individual generators $g$. Each generator consists of an LSTM decoder, initialized with the encoded features $c$ and combined with a random noise vector $ z \sim \mathcal{N}(0,\,1)$ as the initial hidden state $h^{0}$. 
The final trajectory $\hat{Y}$ is predicted recurrently:
\begin{equation}
\Delta \hat{Y}^t = \LSTM_g\left( \Delta \hat{Y}^{t-1}, h^{t-1}\right),
\end{equation}
where $\Delta \hat{Y}^{t_{obs}}$ is initialised with the last displacement of the observation $\Delta X^{t_{obs}}$.

\subsection{Path Mode Network (PM-Net)}
The Path Mode Network $\Pi(c_i)$ outputs a probability $\boldsymbol{\pi}$ over the generators conditioned on the encoded features $c_i$ for a pedestrian $i$. The network consists of a 3-layer Multi-Layer Perceptron (MLP) with ReLU activations and the hidden dimension of size $48$ and computes the final distribution over generators with a Softmax layer.

\subsection{Discriminator and Classifier}

Our model is trained in a GAN framework using a discriminator $D$ and classifier $C$. Both networks use shared weights to encode the scene and trajectory mirroring the generator architecture described in \Cref{sup:sec:architecture:encoding} to obtain the encoding $c$ for a pedestrian.
Additionally, we encode either the predicted or ground-truth trajectory, $Y$ or $\hat{Y}$ respectively, with a two-layer MLP and concatenate with $c$ to obtain the input for the two separate branches of the discriminator $D$ and classifier $C$.
Both use a two-layer MLP and produce the probability of the trajectory being real through a Sigmoid activation (Discriminator) or a distribution from which generator the trajectory was sampled from a Softmax activation (Classifier). Unless otherwise specified, we use LeakyReLU activations with a slope of 0.2 across the module.

\section{Training \modelname{}}
\label{sup:sec:training}
To train \modelname{}, we present an alternating training scheme as is explained in \Cref{supp:TrainingAlgorithm}. The proposed training scheme optimizes the generators and \probnetname{} in the model. During the training, we first optimize \probnetname{} based on the approximated likelihood of the generated trajectories by evaluating $l$ samples each. In the second step, we sample $q$ trajectories from the generator and apply the adversarial loss, best-of-many loss, and classification loss to the trajectories. In this Section, we provide additional information on the derivation of the \probnetname{} training objective (\Cref{sec:supplementary:training:PMobjective}) and discuss the effect of hyperparameters for the training (\Cref{sec:supplementary:training:hyperparameters}).

\subsection{\probnetname{} Objective}
\label{sec:supplementary:training:PMobjective}
To estimate probabilities of trajectories, we assume normal distributed errors of the ground-truth $p_Y = Y + N(0, \sigma I) = N(Y, \sigma I)$. Thus, we can define the probability of a prediction $\hat{Y}$ as $p(\hat{Y} | c, z, g) = p_Y(\hat{Y})$ where $z \sim N(0, I)$ is the GAN noise distribution, $c$ the encoded, conditional scene information and $g$ the generator index.
\\
Using symmetry of the Normal distribution and marginalizing $z$ through $l$ Monte Carlo samples $\{z_{(i)}\}_{i=1}^{l}$, the likelihood of the ground-truth trajectory $Y$ can be written as 
\newcommand{\appropto}{\mathrel{\vcenter{
  \offinterlineskip\halign{\hfil$##$\cr
    \propto\cr\noalign{\kern2pt}\sim\cr\noalign{\kern-2pt}}}}}
\begin{align}
    p(Y | g, c) &= \int p(Y | g, z, c) dz \nonumber \\
                &\approx \frac{1}{l} \sum_{i=1}^{l} p(Y| c, z_{(i)}, g) \nonumber \\
                &\approx \frac{1}{l} \sum_{i=1}^{l} \mathcal{N}(Y; \hat{Y}_{c, z_{(i)}, g}, \sigma I) \nonumber  \\
                &\appropto \frac{1}{l} \sum_{i=1}^{l} \exp \left( \frac{- \norm{\hat{Y}_{g, c, z_i}- Y}_2^2}{2\sigma} \right). \label{supp:eq:likelihood_approx}
\end{align}
\\
By applying Bayes' rule, one obtains the posterior distribution over generators
\begin{align*}
    p(g|Y, c) &= \frac{p(Y|c, g) p(g|c)}{p(Y|c)} \nonumber \\
              &= \frac{p(Y|c, g) p(g|c)}{\sum_g p(Y| g, c)p(g|c)}.
\end{align*}
We use a non-informative, uniform prior distribution over generators $p(g|c) = 1 / n_G$ as we do not have knowledge which generator is relevant for a given scene context $c$ at the start of training. Overall, we obtain
\begin{align}
    p(g|Y, c) &= \frac{p(Y|c, g)}{\sum_h p(Y| h, c)}  \label{sup:eq:pm_net_target}
\end{align}
which can be computed using the approximation of \Cref{supp:eq:likelihood_approx} concluding our derivation.

As described in the main paper, we train \probnetname{} by minimizing the Cross-Entropy between the approximated distribution over generators in \Cref{sup:eq:pm_net_target} which we derived in this section and the output distribution $\Pi(c)$ produced by \probnetname{}.

\begin{algorithm*}[p]
  \caption{Proposed algorithm for training \modelname{}.}  
  \label{supp:TrainingAlgorithm}
  \begin{algorithmic}[1]  
    \Require{$p(z)$ noise distribution, $m$ batch size, $\{\theta_i\}_{i=1}^{n_G}$ set of generator weights, $w$ weights of discriminator, $\zeta$ weights of \probnetname{}, $\lambda_{Traj}$ weighting for $L2$ best-of-many loss, $\lambda_{Cl}$ weighting for generator classification regularization, $q$ number of $G$ training samples, $l$ number of \probnetname{} samples}
    \Repeat
    
    \Sample{$\{x^i\}_{i=1}^m,\{y^i\}_{i=1}^m$}{$p_r(x, y)$} \Comment{Batch from real data where $x$ is the input (observed trajectories and image crop) and $Y$ the ground-truth observation}
    \Sample{$\{z^i\}_{i=1}^m$}{$p(z)$} \Comment{Batch from noise distribution}
    \Sample{$\{c^i\}_{i=1}^m$}{$\Pi(x^i; \zeta)$} \Comment{Batch of generator indices samples from \probnetname{}}
    \Let{$\{\hat{Y}_g^i\}_{i=1}^m$}{$G(x^i, z^i;\theta_{c^i})$} \Comment{Generate batch using selected generators}    
    \Let{$g_w$}{$\nabla_w \frac{1}{m}\sum_i \left[ \ln D(x^i, y^i; w) + \ln (1-D(x^i, \hat{Y}_g^i; w)) \right]$ \label{alg:dmganpl:vd}}
    \Let{$w$}{Adam($w$, $g_w$)} \Comment{Optimize discriminator $D$}
    
    \Let{$g_\gamma$}{$\nabla_\gamma \frac{1}{m} \sum_{i} \left[\ln C(x_g^i, \hat{Y}_g^{i}; \gamma) \right]_{c^i}$} 
    \Let{$\gamma$}{Adam($g_\gamma$, $\gamma$)} \Comment{Update Classifier $C$}
    \\
    \Sample{$\{z^{i,j}\}_{i=1,j=1}^{m,q}$}{$p(z)$}
    \Sample{$\{c^{i,j}\}_{i=1,j=1}^{m,q}$}{$\Pi(x^i; \zeta)$}
    \Let{$\{\hat{Y}_g^{i,j}\}_{i=1,j=1,g=c^{i,j}}^{m,k}$}{$G(x^i,z^{i,j};\theta_{c^{i,j}})$}
    \Let{$\{t^i\}_{i=1}^m$}{$\min_{j, g} \norm{y^i - \hat{Y}_g^{i,j}}$}
    \Let{$\{g_{min}^i\}_{i=1}^m$}{$\arg \min_{g} \norm{y^i - \hat{Y}_g^{i,j}}$}
    \For{$o \in \{ 1 \dots n_G\}$}
        \Let{$g_{\theta_o}$}{$\nabla_{\theta_j} \frac{1}{mk}\sum_{i} \sum_{j} \left[ \ln D(x^i, \hat{Y}_g^{i,j}; w) - \lambda_{Cl} \ln C(x^i, \hat{Y}_g^{i,j}; \gamma) \right] + \delta_{o,g_{min}^i} \lambda_{Traj} t^i$}
        \Let{$\theta_{Cl}$}{Adam($g_{\theta_o}$, $\theta_o$)} \Comment{Optimize generators}
    \EndFor
    \\
    \Sample{$\{z^{i,j}\}_{i=1,j=1}^{m,l}$}{$p(z)$}
    \Let{$\{\hat{Y}_o^{i,j}\}_{i=1,j=1,o=1}^{m,l,n_G}$}{$G(x^i,z^{i,j};\theta_{o})$}
    \Let{$\{p^{i, o}\}_{i=1, o=1}^{m, n_G}$}{$\mathcal{N}(\hat{Y}_o^{i,j}; y^i, \sigma I)$}
    \Let{$g_\zeta$}{$\nabla_\zeta \frac{1}{m} \sum_i H \left( \Pi(x^i; \zeta), p^i) \right)$} 
    \Let{$\zeta$}{Adam($g_\zeta$, $\zeta$)} \Comment{Optimize \probnetname{}}
    \Until{convergence.}
  \end{algorithmic}  
  
\end{algorithm*}

\subsection{Hyperparameters}
\label{sec:supplementary:training:hyperparameters}
For optimization, we use the Adam \cite{adamOptimizer} optimizer with learning rate $0.001$, $\beta_1 = 0.5$ and $\beta_2=0.999$. We set the number of $G$ training samples $q=20$ and \probnetname{} training samples $l=1$. Further, we set the weighting coefficients $\lambda_{Traj},\lambda_{Cl}$, and the standard deviation $\sigma$ to $1$.
We study different settings of the hyperparameters specific to our model on the synthetic dataset in the following.

\paragraph{Effect of Loss Weighting.}
Results for different settings of $\lambda_{Traj}$, weighting the $L2$ best-of-many loss~\cite{best-of-many-sampling, social_gan} loss term $L_{Traj}$, and $\lambda_{Cl}$,  weighting the classifier loss term $L_{Cl}$, can be found in \Cref{supp:tab:loss_weight_ablation}. We observe that higher settings of $\lambda_{Traj}$ help slightly to improve ADE, Precision, and Recall. A higher $\lambda_{Traj}$ enforces that at least one generated sample is close to the ground-truth trajectory reducing ADE and increasing Recall. Additionally, it enforces further specialization of the generators as the $L_{Traj}$ loss only applies to the closest sample that always comes from the same generator when they already cover distinct modes. As a result, we find that $\lambda_{Cl}$ does not affect the results significantly since it encourages the generators to be distinguishable. If the classifier drives the generators to cover distinct modes already in the early stages of training, the loss is small and has little influence during the remaining training independent of the weighting. 

\begin{table}
\centering
\caption{Results of \modelname{} $(n_G = 5)$ trained with different values for $\lambda_{Traj}$ and $\lambda_{Cl}$.}
\label{supp:tab:loss_weight_ablation}
\setlength{\tabcolsep}{1mm}
\begin{tabular}{llrrrr}
\toprule
 $\lambda_{Traj}$ &  $\lambda_{Cl}$ &  ADE &  FDE &  Precision &  Recall \\
\midrule
            0.1 &              0.1 &      0.37 &      0.46 &       0.60 &         0.80 \\
            0.1 &              1.0 &      0.36 &      0.44 &       0.64 &         0.80 \\
            0.1 &              5.0 &      0.34 &      0.44 &       0.68 &         0.82 \\
\midrule
            1.0 &              0.1 &      0.35 &      0.46 &       0.73 &         0.86 \\
            1.0 &              1.0 &      0.33 &      0.44 &       0.71 &         0.90 \\
            1.0 &              5.0 &      0.32 &      0.45 &       0.76 &         0.94 \\
\midrule
            5.0 &              0.1 &      0.32 &      0.46 &       0.80 &         0.92 \\
            5.0 &              1.0 &      0.30 &      0.44 &       0.81 &         0.97 \\
            5.0 &              5.0 &      0.32 &      0.45 &       0.79 &         0.93 \\
\bottomrule
\end{tabular}

\end{table}

\paragraph{Effect of $\sigma$ for \probnetname{} Training.}
The parameter $\sigma$ represents the standard deviation of the normally distributed error  $\mathcal{N}(0, \sigma I)$ in meters around the ground-truth trajectory $Y$ and shapes the likelihood approximation in \Cref{supp:eq:likelihood_approx}. For large $\sigma$ the probability over generators $p(g|Y,c)$ becomes uniform in \Cref{sup:eq:pm_net_target} while the probability converges to a one-hot vector for the generator producing the closest samples to the ground-truth as $\sigma \to 0$.
We find in \Cref{supp:tab:sigma_ablation} that the results of \modelname{} are stable \wrt{} reasonable choices of $\sigma$ indicating that no hyperparameter tuning on $\sigma$ is necessary in order for \modelname{} to converge to a correct solution producing high precision.

\begin{table}
\centering
\caption{Results of \modelname{} $(n_G = 5)$ trained with different values for $\sigma$.}
\label{supp:tab:sigma_ablation}
\begin{tabular}{lrrrrr}
\toprule
$\sigma$ &  ADE &  FDE &  Precision &  Recall \\
\midrule
0.1 &      0.33 &      0.49 &       0.77 &         0.95 \\
0.5 &      0.38 &      0.53 &       0.70 &         0.89 \\
1.0 &      0.32 &      0.44 &   	0.77 &	       0.95 \\	
2.5 &      0.32 &      0.48 &       0.80 &         0.95 \\
5.0 &      0.34 &      0.47 &       0.72 &         0.90 \\
\bottomrule
\end{tabular}
\end{table}

\paragraph{Effect of the Number of Training Samples.}
For the training of \modelname{} we can choose the number of Monte Carlo samples $l$ used for the likelihood estimation in \Cref{supp:eq:likelihood_approx} and the number of generator training samples $q$  for computing the $L2$ best-of-many loss~\cite{best-of-many-sampling, social_gan} $\mathcal{L}_{Traj}$. 
In \Cref{supp:tab:num_samples_ablation}, we find that a single sample $q=1$ is not enough to train a multimodal model, because it still results in a model predicting linear straight motion. By increasing $q$ we find that this model can predict more multimodal trajectories since the loss is only applied on the sample closest to the ground truth and the other $q-1$ predictions in potentially other directions are not punished. Increasing the number of samples $l$ has a positive effect on the performance because the likelihood estimation becomes more accurate and the generators can develop the specialization to a specific mode even further. 

\begin{table}
\centering
\caption{Results of \modelname{} $(n_G = 5)$ trained with different number of \probnetname{} training samples $l$ and generator samples $q$.}\label{Table:num_samples_ablation}
\label{supp:tab:num_samples_ablation}
\setlength{\tabcolsep}{1mm}
\begin{tabular}{llrrrr}
\toprule
\makecell{$\Pi$-Net \\
Samples $l$ }& 
\makecell{$G$ Training \\ Samples $q$}&  ADE &  FDE & Precision &  Recall 5 \\
\midrule
1 &                     1 &      2.18 &      4.57 &       0.31 &         0.30 \\
1 &                    10 &      0.46 &      0.52 &       0.55 &         0.66 \\
1 &                    20 &      0.45 &      0.57 &       0.57 &         0.82 \\
\hline
5 &                     1 &      2.20 &      4.56 &       0.33 &         0.29 \\
5 &                    10 &      0.46 &      0.49 &       0.56 &         0.68 \\
5 &                    20 &      0.31 &      0.45 &       0.79 &         0.95 \\
\bottomrule
\end{tabular}

\end{table}

\section{Definition of Precision and Recall}
\label{sec:precisionrecall}
To measure the performance of models in preventing out-of-distribution (OOD) samples while covering the entire support of the distribution, we follow the GAN literature \cite{FirstPrecisionRecallNIPS, ImprovedPrecisionRecallNIPS}  and estimate the manifolds of predicted trajectories and ground-truth samples to compute Precision and Recall. 

\begin{figure}
  
    \begin{subfigure}[t]{0.155\textwidth}
      \centering
        \includegraphics[width=\textwidth]{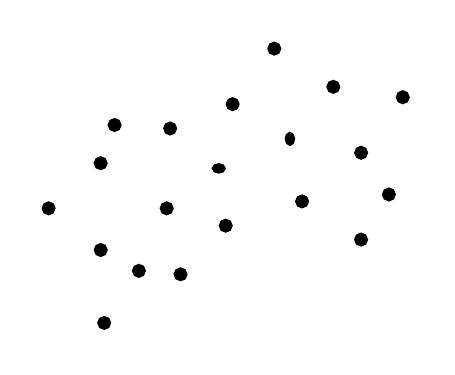}
        \caption{}
        \label{fig:metric:points}
    \end{subfigure}
    \hfill
    \begin{subfigure}[t]{0.155\textwidth}
      \centering
        \includegraphics[width=\textwidth]{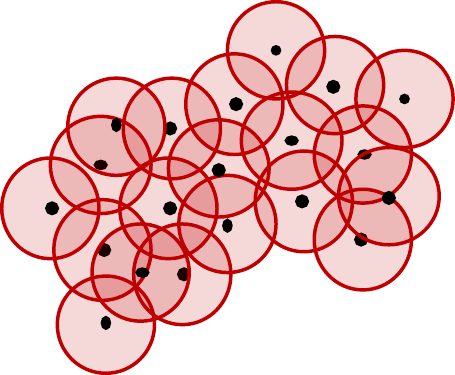}
        \caption{}
        \label{fig:metric:radius}
    \end{subfigure}
    \hfill
      \begin{subfigure}[t]{0.155\textwidth}
        \centering
        \includegraphics[width=\textwidth]{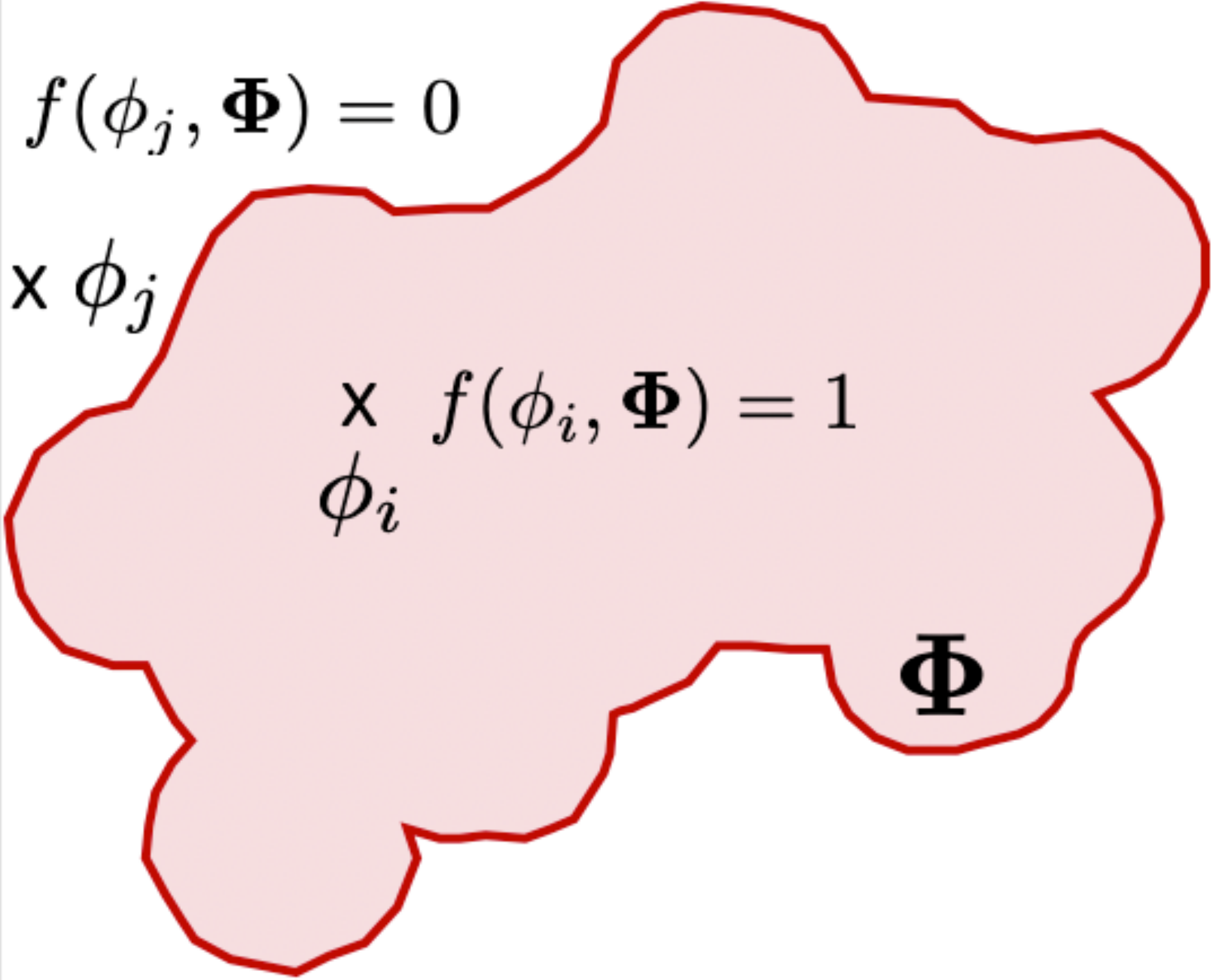}
        \caption{}
        \label{fig:metric:manifold}
    \end{subfigure}
    \caption{(a) Trajectory endpoints, (b) estimating manifold through disc around the points with radius $R$, and (c) testing if samples $\phi$  are in estimated manifold. }
    \label{fig:Metrics:RecallPrecision}
\end{figure}

For a set of future trajectories $\Phi = \{\phi_k\}$, we estimate the corresponding manifold in the output space by considering the points of all trajectories $\{\phi^t_k\}$ at time $t$ (\Cref{fig:metric:points}), constructing a disc with radius $R^t$ around each point $\phi^t_k$  (\Cref{fig:metric:radius}). The $R^t$ represents the maximum distance error we can accept for the predictions. The union of all disc areas serves as an estimate of the true manifold (\Cref{fig:metric:manifold}). We do this for every time $t \in \{1, ..., T\}$ and define 
\begin{equation*}
    R^t = \frac{R_{max} \cdot t}{T}
\end{equation*} where we set $R_{max} = 2m$.
To determine if a given sample $\phi$ lies inside this manifold, we define a binary score function:
\begin{equation}
score(\phi, \boldsymbol{\Phi})=\begin{cases}
1, \text{ if } \forall t \exists \phi^{\prime} \in \boldsymbol{\Phi} \text{ with } \norm{\phi_t - \phi_t^\prime}_2 \leq R^t  \\
0, \text{ otherwise.}
\end{cases}
\label{eq:manifold_estimate}
\end{equation}
Following the above definition, we construct the manifold $\boldsymbol{\Phi}_G$ based on the set of model predictions $\boldsymbol{\phi}_G$. Similarly, we estimate the ground-truth manifold $\boldsymbol{\Phi}_R$ using the set of ground-truth trajectories $\boldsymbol{\phi}_R$ as provided in the FPD~\cite{liang2020garden} and the synthetic dataset (\Cref{supp:section:synthetic_datset}). 

Precision and Recall are then defined using \Cref{eq:manifold_estimate} as
\begin{align}
\text{Precision} &= \frac{1}{|\boldsymbol{\phi}_G|} \sum_{\phi \in \boldsymbol{\phi}_G} score(\phi, \boldsymbol{\Phi}_R) \;  \text{and} \;  \\ 
\text{Recall} &= \frac{1}{|\boldsymbol{\phi}_R|} \sum_{\phi \in \boldsymbol{\phi}_R} score(\phi, \boldsymbol{\Phi}_G). \label{eq:knn_precision_recall}
\end{align}

Intuitively, Precision measures the realism of a trajectory because it queries if the prediction falls inside the ground-truth manifold. Symmetrically, Recall measures if real samples lie within the manifold generated by predictions and thus measures if all modes present in the ground truth are covered.

\section{Synthetic Dataset}
\label{supp:section:synthetic_datset}
\begin{figure}
\begin{center}
       \begin{subfigure}[t]{0.49\linewidth}
       \centering
       \includegraphics[width=0.7\linewidth]{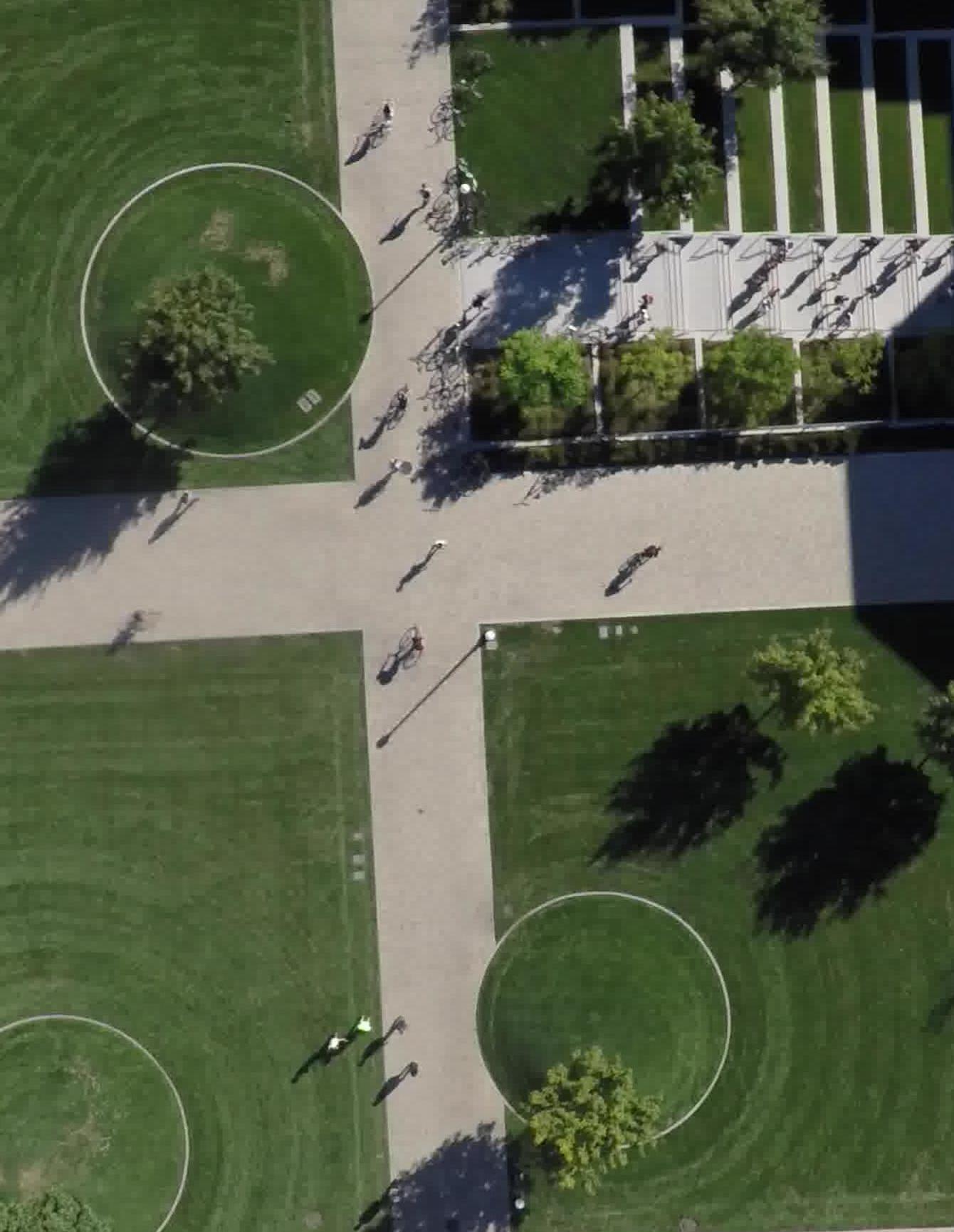}
       \caption{}
       \label{fig:synthetic:hyang}
       \end{subfigure}
         \hfill
      \begin{subfigure}[t]{0.49\linewidth}
      \centering
      \includegraphics[width=0.7\linewidth]{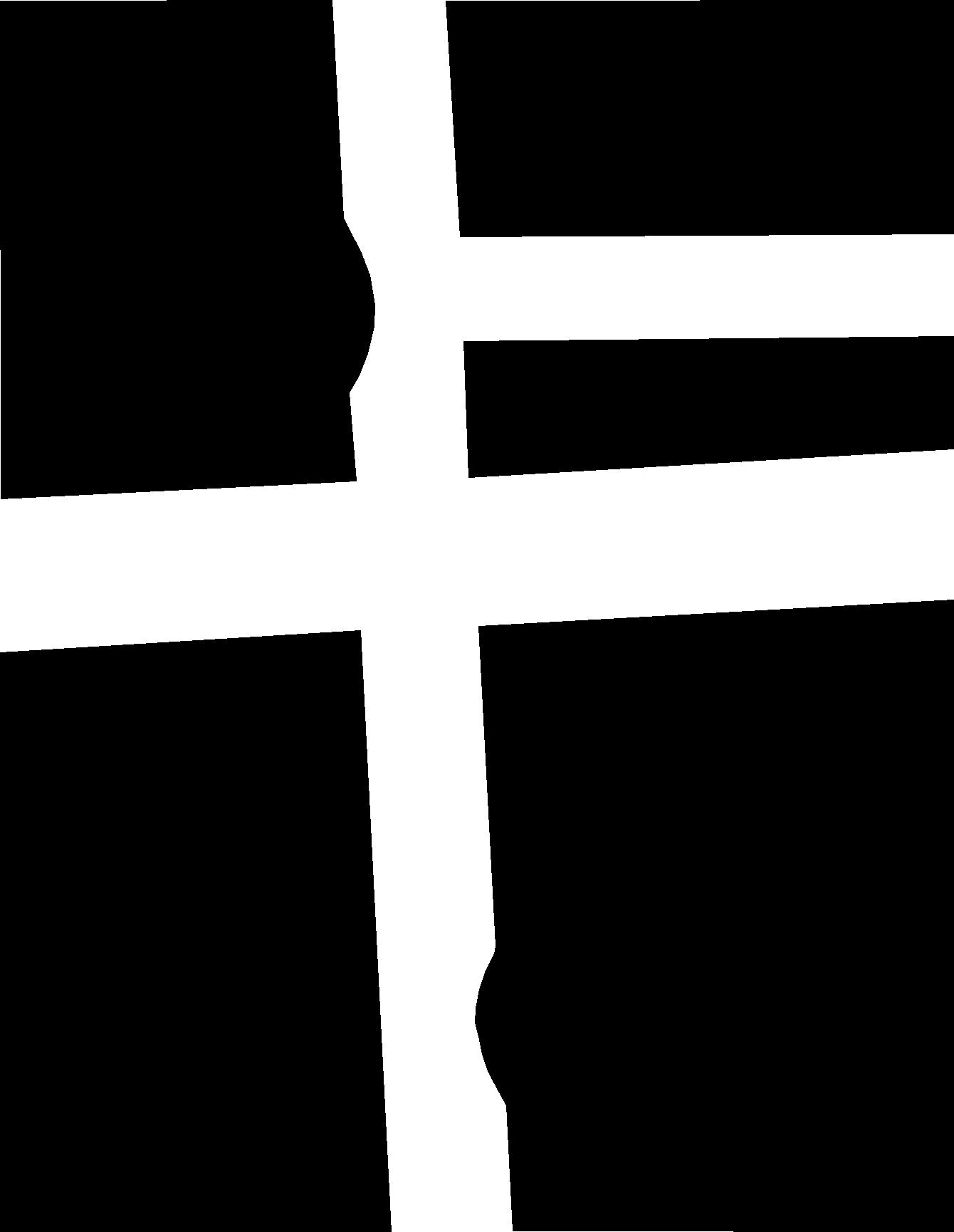}
      \caption{}
      \label{fig:synthetic:seg}
      \end{subfigure}
     \caption{Scene image (a) and occupation map (b) of the synthetic dataset.}
     \label{fig:synthetic:dataset}
\end{center}
\end{figure}

In the main paper, we present a synthetic dataset to study the generated multimodality of the models. We generate the dataset on the Hyang-4 scene of the SDD~\cite{stanforddronedataset}, as shown in \Cref{fig:synthetic:hyang}. This scene is well suited because it provides separated spatial modes with an upper and lower junction with two and three modes respectively.
We simulate the dynamics of $\approx 80,000$ pedestrians using the Social Force Model~\cite{social_force}. In order to control and limit the modes of future trajectories, we use an occupancy map shown in \Cref{fig:synthetic:seg} restricting the area the pedestrians can walk on. In the dataset, we primarily focus on spatial multimodality and limit the number of pedestrians to a maximum of two.

\section{Multimodality of Real Datasets}
\label{sup:sec:multimodality_real_datasets}
In this section, we try to measure the overall multimodality of or public benchmarks.
 As we do not have multiple ground-truth trajectories on these datasets, we consider similar trajectories which are (i) in close proximity (closer than $2m$), (ii) walk-in similar directions ($ \pm 45^{\circ} $), and (iii) walk with similar speed ($ \pm 0.5 \nicefrac{m}{s} $). We then filter trajectories if they collide with other pedestrians in the scene (distance $\leq 0.5m$). 

Finally, we use the procedure described in \Cref{sec:precisionrecall} to estimate the manifold based on the collected set of trajectories and count the number of disconnected components across timesteps. 

\begin{figure}
    \centering
    \includegraphics[width=0.7\linewidth]{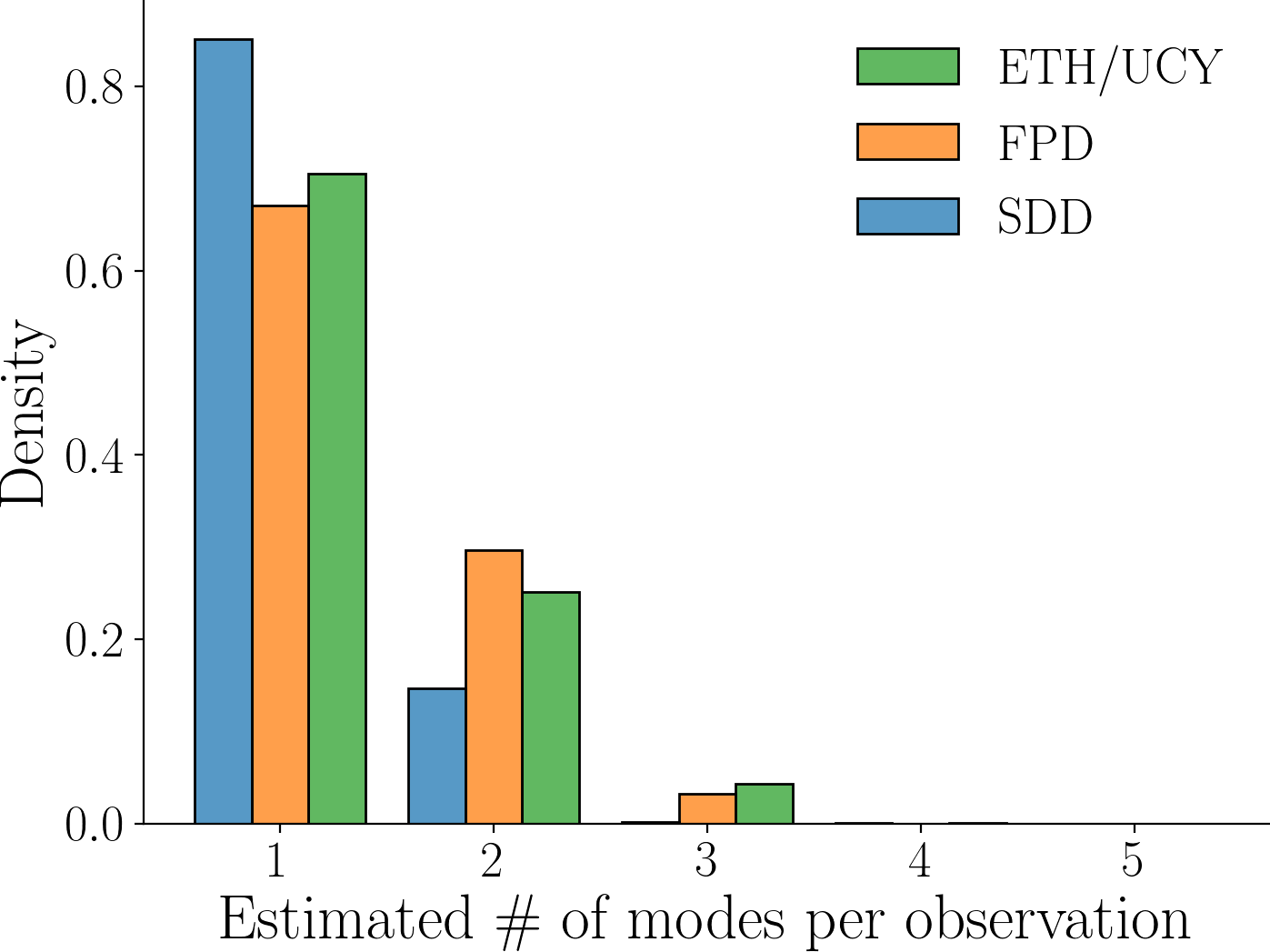}
    \caption{Histogram of the estimated, relative number of modes for the real datasets.}
    \label{supp:fig:estimated_number_modes}
\end{figure}

We conclude from \Cref{supp:fig:estimated_number_modes} that SDD is less multimodal (Avg. \# of modes: $1.15$) than the other datasets, i.e., FPD (Avg. \# of modes: $1.36$) and UCY/ETH (Avg. \# of modes: $1.34$). This observation is somewhat congruent with the performance of \modelname{} on the public benchmarks. While our method achieves state-of-the-art performance on ETH and UCY which is more multimodal than SDD that naturally makes it hard for our method to show benefits over existing methods.  

We elaborate this further and train the single generator baseline \textit{GAN+L2} with the same backbone on SDD. We obtain similar results compared to \modelname{} with ADE of $14.6$ and FDE $27.5$ as shown in \Cref{Table:sdd_results_num_gnerators_supp}. These results further indicate that SDD does not contain sufficient multi-modality for our method achieving better results than single-generator methods.
 
\begin{table}
\centering
\caption{ADE and FDE results for \modelname{} with different number of generators on the SDD dataset.}\label{Table:sdd_results_num_gnerators_supp}
\begin{tabular}{lrrrr}
\toprule
\# Generators &     2 &     3 &     4 &     5 \\
\midrule
ADE &  13.7 &  14.6 &  13.6 &  14.5 \\
FDE &  26.1 &  27.6 &  25.8 &  27.6 \\
\bottomrule
\end{tabular}

\end{table}

\section{Toy Experiment} 
\label{sup:sec:toy_experiment}
In addition to the experiments in the main paper, we study multimodality on a toy dataset introduced in \cite{social_ways}. The data consists of six starting positions equidistantly distributed on a circle where we generate three paths with uniform probability for each starting position, as shown in \Cref{fig:socialwayssyntehtic_gt}. This experiment demonstrates how different models represent the multiple modes inside a lower-dimensional latent space. This experiment should give us a deeper understanding of the architectural requirements for modeling distinct modes.
For this experiment, all methods use a single encoder to encode the observation, then an MLP is used to transform the encoding with the GAN noise $z$ to the three-dimensional latent space, and a decoder decodes the latent space to predictions. For \modelname{}, we use separate MLPs mapping to the latent space resembling the different generators in our main model. 

\paragraph{Results.} 
In \Cref{fig:exp_latent_toy_single}, we visualize the predictions (left) and corresponding latent space vectors (right) for the considered models.
The simple GAN baseline fails to recover all three modes, while training with an additional $L2$ loss leads to unrealistic out-of-distribution samples since the loss encourages the samples to spread over the entire output space which is also reflected in the latent space. 
InfoGAN does not learn to encode different modes inside its categorical values. Ultimately, our multi-generator model covers all disconnected modes without producing out-of-distribution samples in between. The different sub-networks learn to map the different modes in separated areas in the latent space and hence introduce the required disconnectedness that allows the prediction of disjoint manifolds. This further demonstrates the efficacy of multi-generator models compared to single generator models in preventing OOD samples while covering the entire distribution. 

\newcommand{\widthImg}{0.40}
\begin{figure*}
    \centering
    \begin{subfigure}[b]{.09\textwidth}
        \includegraphics[width=\linewidth]{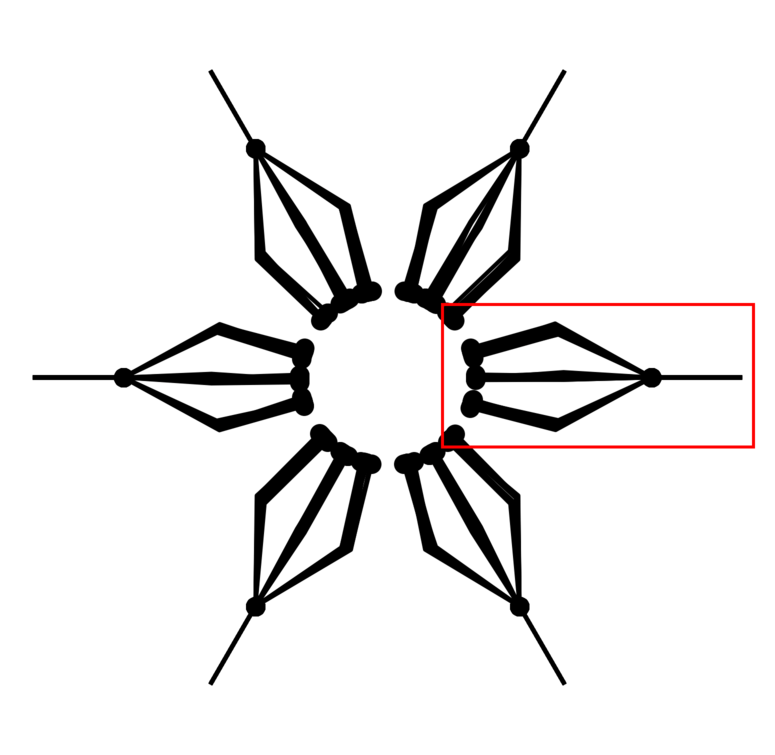}
         \caption{GT}
        \label{fig:socialwayssyntehtic_gt} 
    \end{subfigure}\hfill
    \begin{subfigure}[b]{.22\textwidth}
    \centering
     \raisebox{-0.5\height}{\includegraphics[width=\widthImg\textwidth]{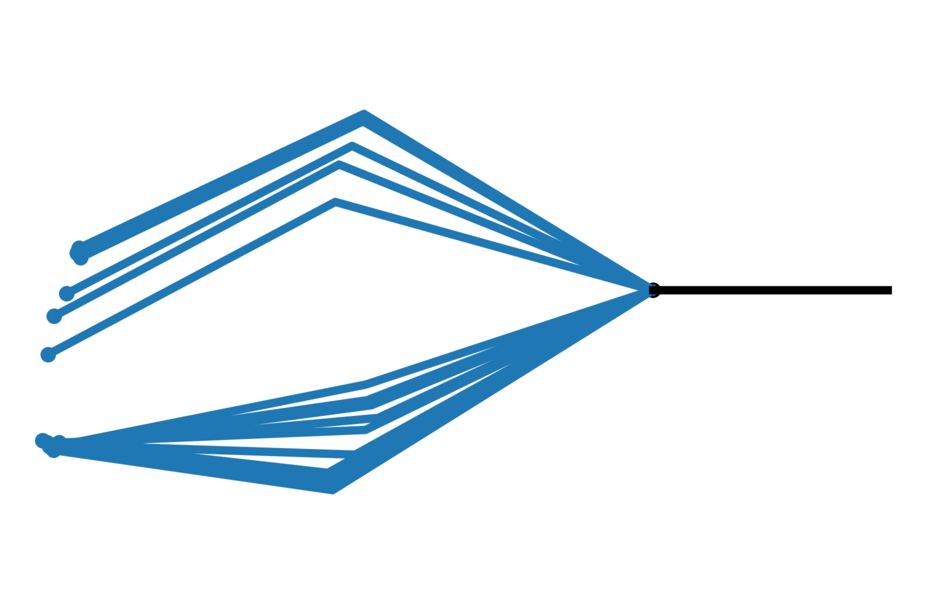}}
    \raisebox{-0.5\height}{\includegraphics[width=\widthImg\textwidth]{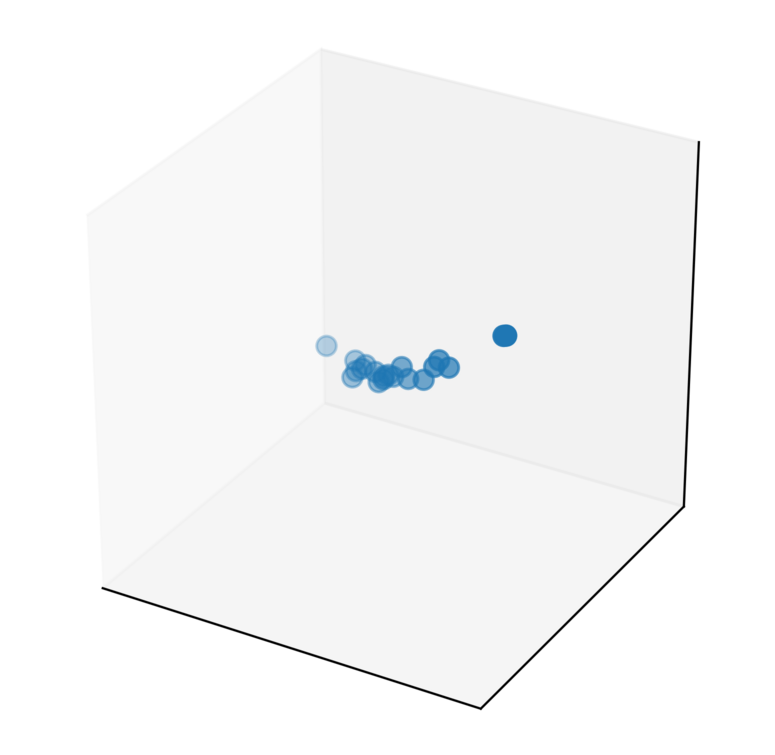}}
        \caption{GAN}
    \end{subfigure}\hfill
     \begin{subfigure}[b]{.22\textwidth}
    \centering
    \raisebox{-0.5\height}{ \includegraphics[width=\widthImg\textwidth]{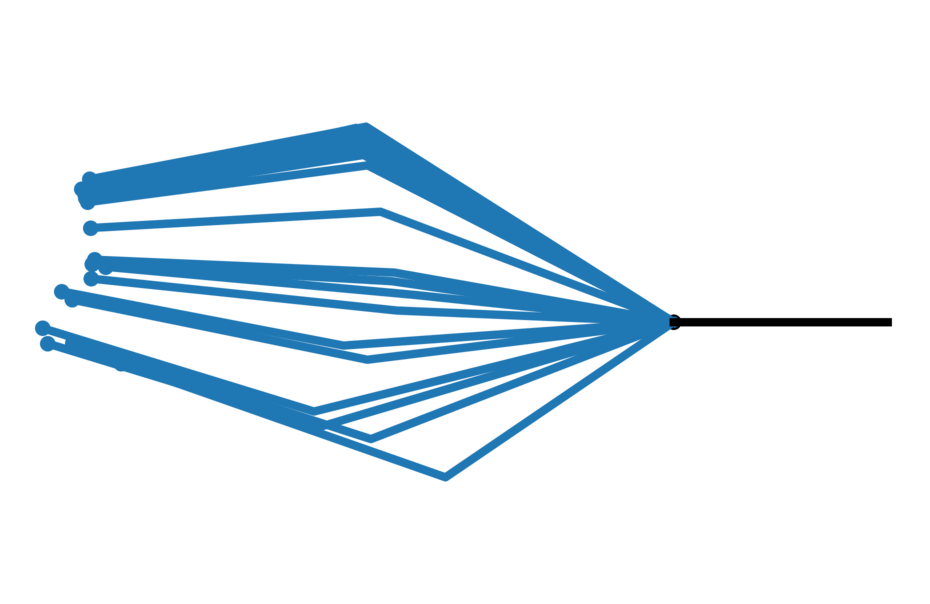}}
   \raisebox{-0.5\height}{ \includegraphics[width=\widthImg\textwidth]{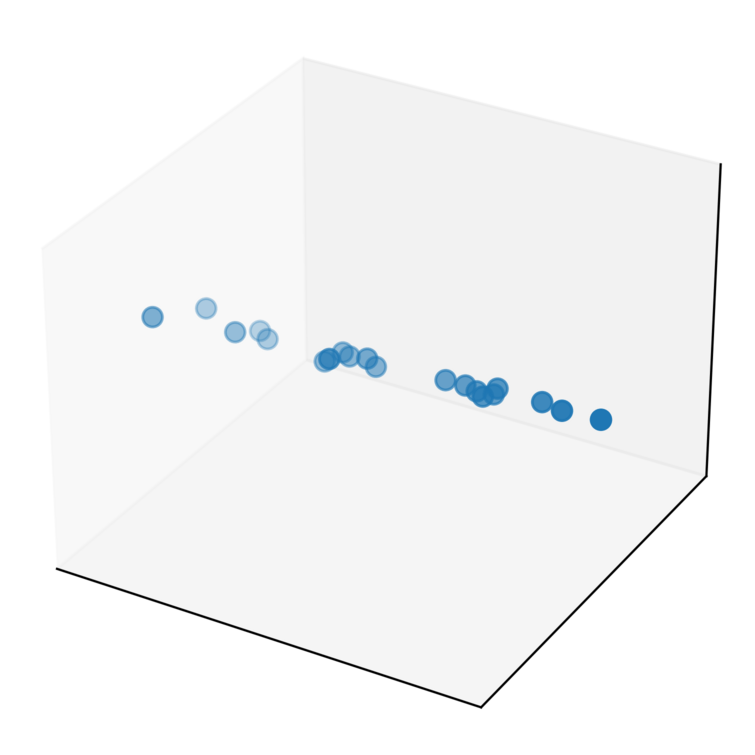}}
        \caption{GAN+L2}
    \end{subfigure}\hfill
    \begin{subfigure}[b]{.22\textwidth}
    \centering
   \raisebox{-0.5\height}{ \includegraphics[width=\widthImg\textwidth]{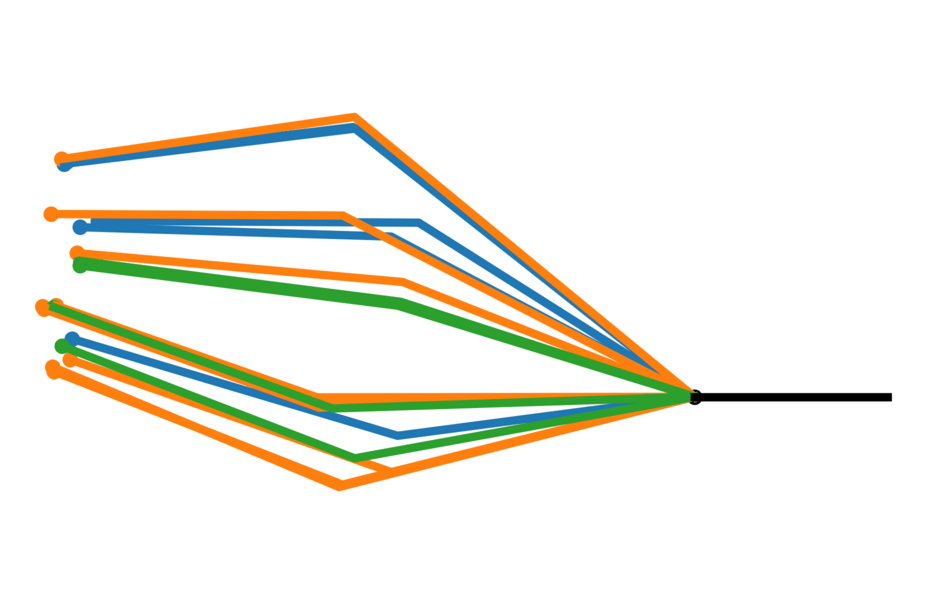}}
  \raisebox{-0.5\height}{  \includegraphics[width=\widthImg\textwidth]{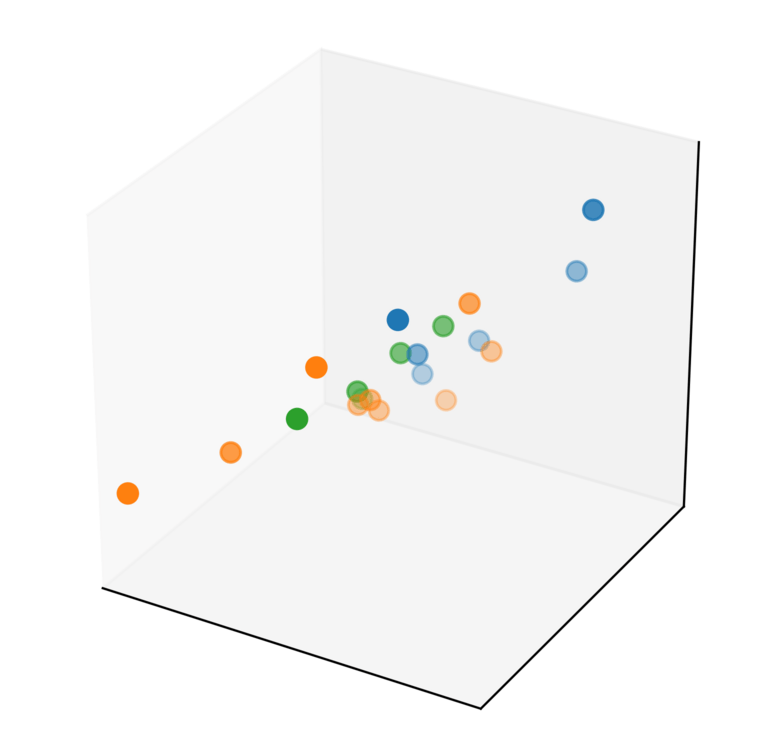}}
        \caption{InfoGAN }
    \end{subfigure}\hfill
    \begin{subfigure}[b]{.22\textwidth}
    \centering
  \raisebox{-0.5\height}{  \includegraphics[width=\widthImg\textwidth]{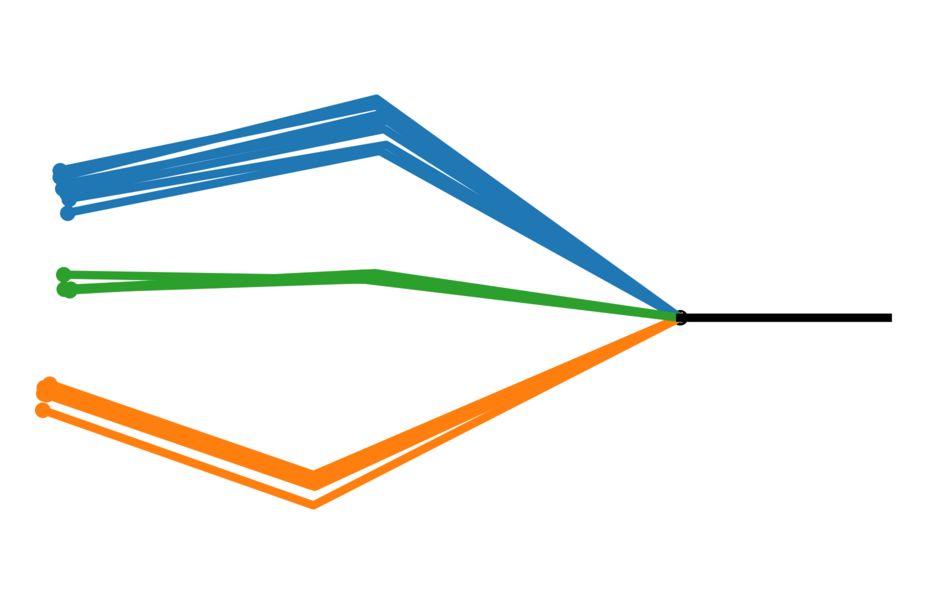}}
  \raisebox{-0.5\height}{  \includegraphics[width=\widthImg\textwidth]{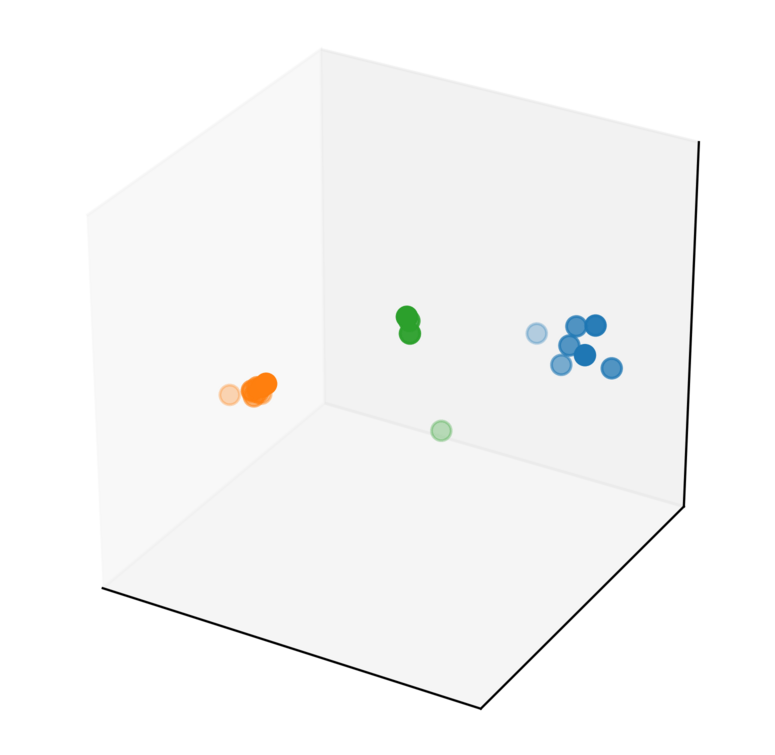}}
        \caption{\modelname{}}
    \end{subfigure}
    \caption{(a) shows the entire toy dataset.  Plots (b)-(e) demonstrate predictions for different models and the corresponding points in the latent space for the observation in the red box (a). Samples from different discrete latent codes or generators are visualized in different colors.}
    \label{fig:exp_latent_toy_single}
\end{figure*}

\section{Visualizations}
\label{sup:sec:visualizations}
\begin{figure*}
\begin{subfigure}[b]{0.3\textwidth}
    \begin{subfigure}[p]{0.45\textwidth}
        \includegraphics[width=\linewidth]{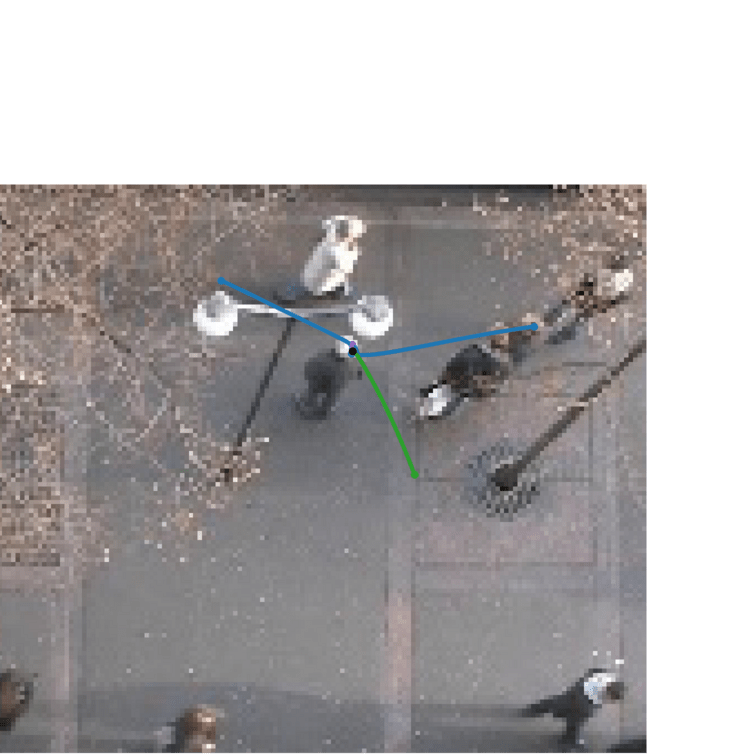}
    \end{subfigure} \hfill
    \begin{subfigure}[p]{0.45\textwidth}
        \includegraphics[width=\linewidth]{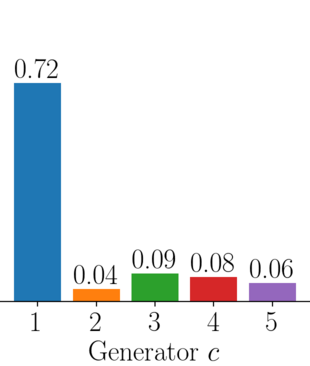}
    \end{subfigure}
\end{subfigure}
\hfill
\begin{subfigure}[b]{0.3\textwidth}
    \begin{subfigure}[p]{0.45\textwidth}
        \includegraphics[width=\linewidth]{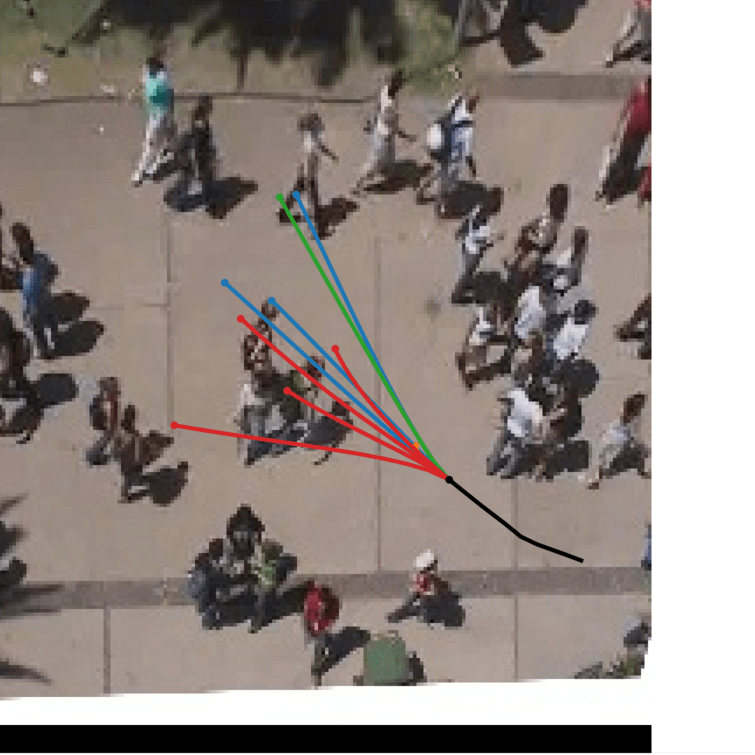}
    \end{subfigure} \hfill
    \begin{subfigure}[p]{0.45\textwidth}
        \includegraphics[width=\linewidth]{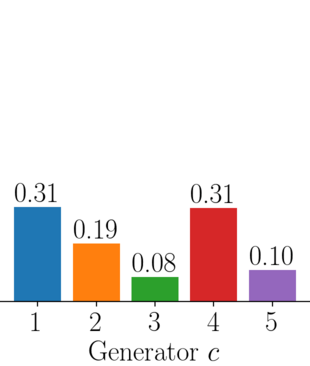}
    \end{subfigure}
\end{subfigure}
\hfill
\begin{subfigure}[b]{0.3\textwidth}
    \begin{subfigure}[p]{0.45\textwidth}
        \includegraphics[width=\linewidth]{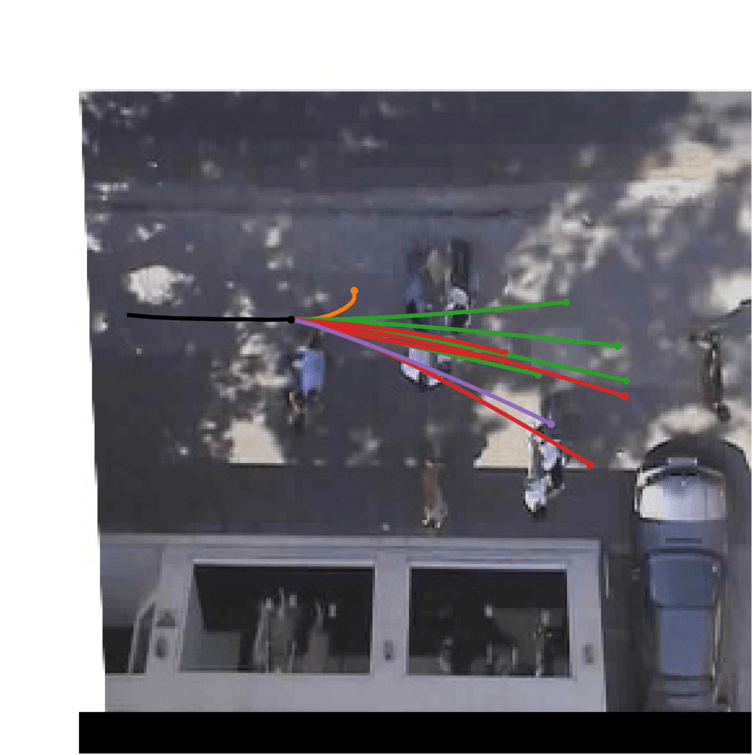}
    \end{subfigure}
    \begin{subfigure}[p]{0.45\textwidth}
        \includegraphics[width=\linewidth]{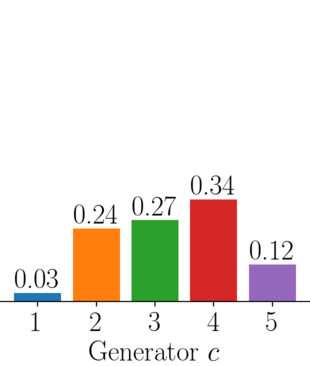}
    \end{subfigure}
\end{subfigure}
\vfill
\caption{Generated samples of \modelname{} on the ETH and UCY dataset}
\label{fig:biwi_samples}
\end{figure*}

\begin{figure*}
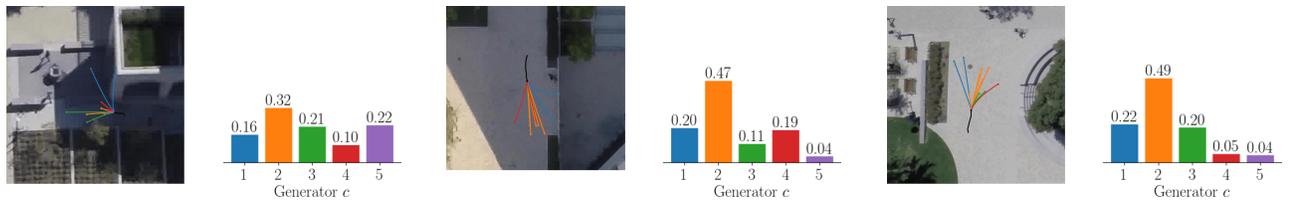


\foreach \x in {1594_4,2457_0,3187_0}
{ 
    \begin{subfigure}[b]{0.3\textwidth}
        \begin{subfigure}[p]{0.45\textwidth}
            \includegraphics[width=\linewidth]{images/sdd_samples/\x.png}
        \end{subfigure} \hfill
        \begin{subfigure}[p]{0.45\textwidth}
            \includegraphics[width=\linewidth]{images/sdd_samples/\x_hist.png}
        \end{subfigure}
    \end{subfigure}
    \hfill
}
\caption{Generated samples of \modelname{} on the Stanford Drone Dataset (SDD)}
\label{fig:sdd_samples}
\end{figure*}

\begin{figure*}
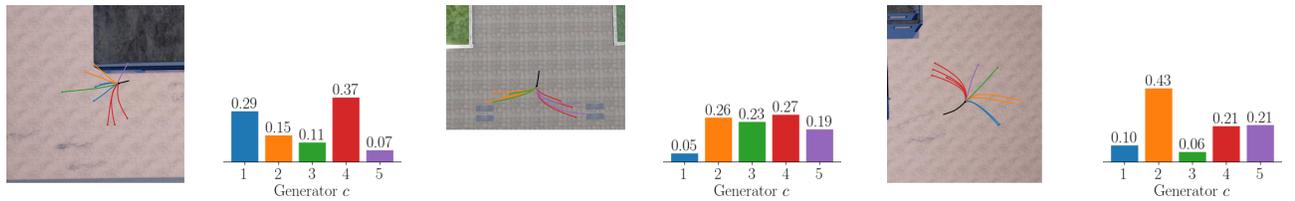


\foreach \x in {3416_0,494_0,5878_4}
{ 
    \begin{subfigure}[b]{0.3\textwidth}
        \begin{subfigure}[p]{0.45\textwidth}
            \includegraphics[width=\linewidth]{images/gofp_samples/\x.png}
        \end{subfigure} \hfill
        \begin{subfigure}[p]{0.45\textwidth}
            \includegraphics[width=\linewidth]{images/gofp_samples/\x_hist.png}
        \end{subfigure}
    \end{subfigure}
    \hfill
}
\caption{Generated samples of \modelname{} on the Forking Path dataset}
\label{fig:gofp_samples}
\end{figure*}

We present additional visualizations of generated trajectories of \modelname{} on ETH and UCY in \Cref{fig:biwi_samples}, SDD in \Cref{fig:sdd_samples} and the FPD in \Cref{fig:gofp_samples}. 

\begin{figure*}

\foreach \x/\y in {single_gan/GAN+L2,multi_gen_0/MG-GAN \newline Generator 1,multi_gen_1/MG-GAN \newline Generator 2,multi_gen_2/MG-GAN\newline  Generator 3, multi_gen_3/MG-GAN \newline Generator 4,multi_gen_4/MG-GAN \newline Generator 5}
{ 
    \subcaptionbox{\y\label{\x}}%
      [0.15\textwidth]{\includegraphics[width=\linewidth]{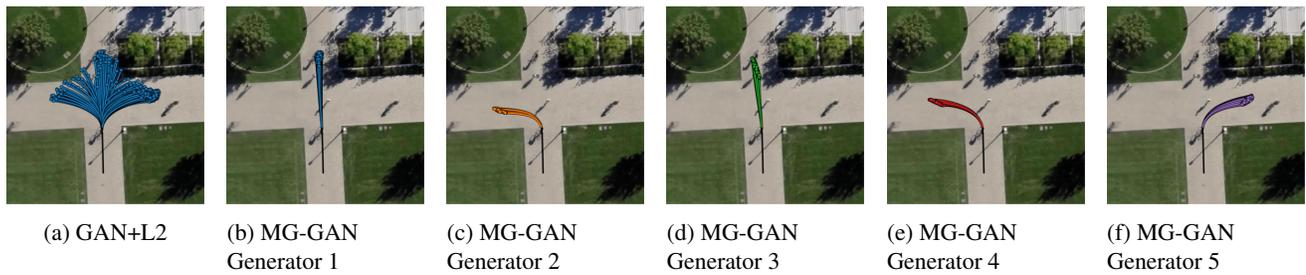}}
    \hfill
}
\caption{Trajectory samples during latent space walk for the single generator model in \Cref{single_gan} and the individual generators of \modelname{} (5) in \Cref{multi_gen_0,multi_gen_1,multi_gen_2,multi_gen_3,multi_gen_4}.}
\label{fig:latent_interpolation}
\end{figure*}

In \Cref{fig:latent_interpolation}, we explore the latent space of the GAN baseline and \modelname{}. While a latent space interpolation results in out-of-distribution samples for the baseline, we show that each generator is limited to the support of a specific mode preventing the generation of OOD samples.

\end{appendices}
\end{document}